\documentclass[10pt]{article}

\iftrue 
  \usepackage[preprint]{tmlr}
  \newcommand{\codeurl}{at \url{https://github.com/maorfarid1/three-clocks}}
  \newcommand{\pdfauthorname}{Maor Farid}
\else
  \usepackage{tmlr}
  \newcommand{\codeurl}{in the anonymized supplementary material}
  \newcommand{\pdfauthorname}{Anonymous}
\fi

\usepackage{amsmath,amsfonts,bm}

\def\eqref#1{equation~\ref{#1}}

\def\1{\bm{1}}

\DeclareMathAlphabet{\mathsfit}{\encodingdefault}{\sfdefault}{m}{sl}
\SetMathAlphabet{\mathsfit}{bold}{\encodingdefault}{\sfdefault}{bx}{n}

\usepackage[most]{tcolorbox}
\usepackage{hyperref}
\hypersetup{colorlinks=true,linkcolor=blue!50!black,citecolor=blue!50!black,urlcolor=blue!50!black}
\hypersetup{pdftitle={Large Language Models as Nonlinear Dynamical Systems: The Three Clocks of Depth, Training, and Generation},
  pdfauthor={\pdfauthorname},
  pdfsubject={Large language models and transformers studied with nonlinear dynamics},
  pdfkeywords={large language models, transformers, nonlinear dynamics, dynamical systems, edge of stability, loss spikes, repetition loops, sampling temperature, Kramers escape, Arrhenius law, Lyapunov exponents, bifurcation, synchronization, Kuramoto model, reduced-order models}}
\usepackage{url}
\usepackage{graphicx}
\usepackage{booktabs}
\usepackage{tabularx}
\usepackage{multirow}
\usepackage{array}
\usepackage{amsmath,amssymb,amsthm}
\usepackage{xcolor}
\usepackage{enumitem}
\usepackage{tikz}
\usepackage{placeins}
\usepackage{longtable}
\usetikzlibrary{arrows.meta,positioning,calc,fit,backgrounds,decorations.pathreplacing,decorations.pathmorphing,patterns}

\newtheorem{proposition}{Proposition}

\theoremstyle{definition}
\newtheorem{definition}{Definition}
\theoremstyle{remark}

\newcommand{\op}[1]{\textbf{OP\,#1}}
\newcommand{\statusP}{\textsf{P}}
\newcommand{\statusM}{\textsf{M}}
\newcommand{\statusC}{\textsf{C}}
\newcommand{\na}{\textsf{?}}

\title{Large Language Models as Nonlinear Dynamical Systems:\\ The Three Clocks of Depth, Training, and Generation}

\author{\name Maor Farid \email maor@getleo.ai \\
      \addr Leo AI, Inc.\\
      50 Milk Street, Boston, MA 02109, United States}

\def\month{MM}
\def\year{YYYY}
\def\openreview{\url{https://openreview.net/forum?id=XXXX}}

\begin{document}

\maketitle

\begin{abstract}
Language models are dynamical systems. A transformer is iterated over its layers, over the steps of its training and over the tokens it generates, and on each of these three clocks it shows behavior that nonlinear dynamics has studied for a century: fixed points, bifurcations, synchronization, chaos and noise-driven escape from potential wells. The two communities rarely meet, so this behavior is often rediscovered under new names. We build a bridge between them. To describe, we condense about 200 papers into the Three-Clock Dictionary, which maps fifteen concepts of nonlinear dynamics onto each clock and marks every correspondence as proved, measured or conjectured, and we show each concept in a mechanical system and in a language model side by side. To explain, we derive reduced-order models: the softmax is a Boltzmann distribution, gradient descent an Euler step on a nonlinear spring, a loss spike a ball thrown out of a stiff valley, attention on the circle a Kuramoto model, and a repetition loop a well that deepens while it is occupied. We then test these explanations on open language models, on a laptop CPU. Escape from repetition loops in GPT-2 and Pythia follows the Arrhenius law with the logit gap as the barrier, and the barrier deepens and then saturates, so the temperature needed to leave a loop falls like $1/\ln N$ with the generation length $N$ and rises with model size. Coherent text lives in a narrow window of temperatures, which a small reasoning model shares at its recommended temperature. Training makes each Pythia layer more expansive, and a small transformer trains at a multi-mode edge of stability. The third step, solving problems of language models, is still ahead: controllers derived from the escape analysis remove loops but do not yet beat the best fixed temperature. We state twelve open problems where nonlinear dynamics could get there.

\end{abstract}

\section{Introduction}
\label{sec:intro}

\begin{figure}[t]
\centering
\begin{tikzpicture}[
  font=\small,
  layer/.style={draw=black!70, fill=blue!8, rounded corners=1.5pt, minimum width=2.3cm, minimum height=0.42cm, inner sep=1pt, font=\scriptsize},
  mini/.style={draw=black!60, fill=green!10, rounded corners=1pt, minimum width=0.85cm, minimum height=0.2cm, inner sep=0pt},
  tokbox/.style={draw=black!60, fill=orange!12, rounded corners=1pt, minimum width=0.62cm, minimum height=0.42cm, inner sep=1pt, font=\scriptsize},
  clock/.style={-{Latex[length=2.4mm]}, line width=1.1pt},
  thin/.style={-{Latex[length=1.5mm]}, black!55},
  note/.style={font=\scriptsize, align=left},
  ttl/.style={font=\small\bfseries, align=left}
]
\foreach \k/\x/\lab/\op in {0/-7.2/{$\theta_0$}/0.35, 1/-6.0/{$\theta_t$}/0.65, 2/-4.8/{$\theta_{\mathrm{end}}$}/1.0} {
  \begin{scope}[opacity=\op]
    \foreach \i in {0,...,5} { \node[mini] at (\x, 0.42+0.27*\i) {}; }
  \end{scope}
  \node[font=\scriptsize] at (\x, 0.05) {\lab};
}
\draw[clock, green!45!black] (-7.75, -0.3) -- (-4.25, -0.3);
\node[ttl, text=green!40!black, anchor=north west] at (-7.8, -0.45) {Clock 2: training time $t$};
\node[note, anchor=north west, text width=4.0cm] at (-7.8, -0.95) {$\theta_{t+1} = \theta_t - \eta\, g_t(\theta_t)$\\[1pt]edge of stability, catapults and loss spikes, saddle plateaus, grokking};
\draw[thin, dashed] (-4.25, 1.1) -- node[above, font=\scriptsize] {weights $\theta$} (-1.3, 1.1);

\foreach \i/\lab in {0/{embed $x^{0}$}, 1/{block $\ell=1$}, 2/{block $\ell=2$}, 3/{$\cdots$}, 4/{block $\ell=L$}} {
  \node[layer] (L\i) at (0, 0.55*\i) {\lab};
}
\node[draw=black!70, fill=gray!12, rounded corners=1.5pt, minimum width=2.3cm, minimum height=0.42cm, inner sep=1pt, font=\scriptsize] (head) at (0, 2.95) {$\mathrm{softmax}(z/T)$};
\draw[thin] (L4.north) -- (head.south);
\draw[clock, blue!70!black] (1.45, -0.2) -- (1.45, 2.45);
\node[ttl, text=blue!60!black, anchor=north west] at (1.65, 2.55) {Clock 1: layer time $\ell$};
\node[note, anchor=north west, text width=3.9cm] at (1.65, 2.05) {$x^{\ell+1}_i = x^{\ell}_i + f_{\theta}^{\ell}\big(x^{\ell}_i;\, x^{\ell}_{1:n}\big)$\\[1pt]tokens as interacting particles; clustering, rank collapse, order to chaos, dynamical isometry};

\foreach \j/\t in {0/{$y_1$}, 1/{$y_2$}, 2/{$\cdots$}, 3/{$y_{n}$}} {
  \node[tokbox] (T\j) at (-1.05+0.7*\j, -1.0) {\t};
}
\node[tokbox, fill=orange!40] (Tn) at (1.85, -1.0) {$y_{n+1}$};
\draw[thin] (0, -0.77) -- (L0.south);
\draw[clock, orange!80!black, rounded corners=8pt] (head.east) -- (6.0, 2.95) -- (6.0, -1.0) -- (Tn.east);
\node[ttl, text=orange!70!black, anchor=north west] at (-1.4, -1.4) {Clock 3: generation time $n$};
\node[note, anchor=north west, text width=7.2cm] at (-1.4, -1.9) {$y_{n+1} \sim \mathrm{softmax}\big(z(y_{1:n};\theta)/T\big)$: repetition loops as attractors, temperature transitions, drift, collapse};
\end{tikzpicture}
\caption{The three clocks of a transformer language model. \textbf{Clock 1} (blue) iterates the residual update over layers for a fixed input and fixed weights. \textbf{Clock 2} (green) iterates the optimizer over the weights; it hands its current weights to the model. \textbf{Clock 3} (orange) iterates sampling: each new token is appended to the context and the whole stack runs again. One tick of clock~3 contains a full run of clock~1, and clock~2 changes the map that both of them iterate. Typical behavior studied on each clock is listed next to it; Table~\ref{tab:dictionary} maps it onto dynamical-systems concepts.}
\label{fig:clocks}
\end{figure}
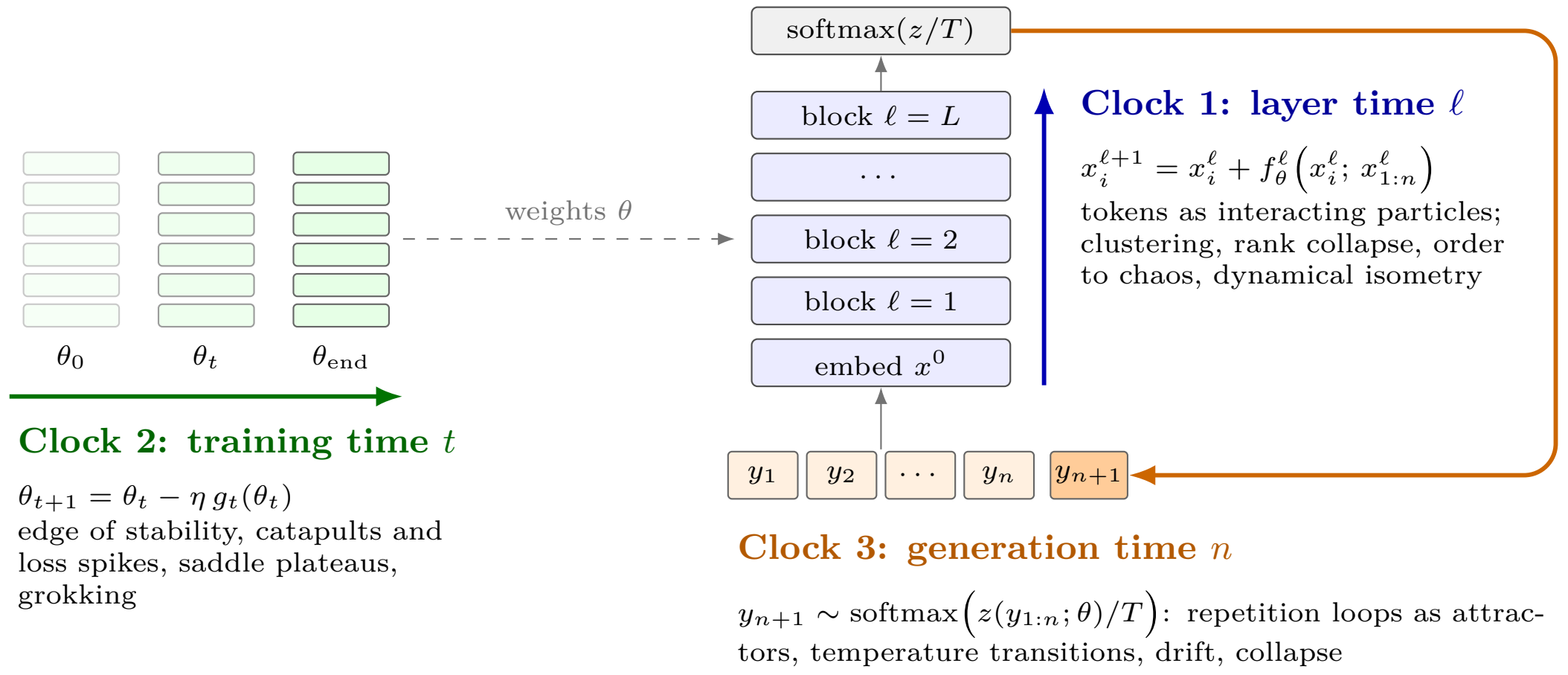

\subsection{Language models, dynamical systems and the potential between them}
\label{sec:potential}

A transformer language model is a nonlinear map. It takes a context, a sequence of tokens, and returns a probability distribution over the next token, much as a forced pendulum maps its initial condition and excitation to a history of displacements.\footnote{Use of AI tools: an AI assistant (Claude, Anthropic) helped search and verify the literature, draft and edit text, and write and run the experiment code, under the author's direction. All references were checked against arXiv or Crossref records and all reported numbers come from the released code. The author takes full responsibility for the content.} This analogy has one important limit. In a mathematical model of a physical or mechanical system, the parameters, such as the length of a pendulum or the stiffness of a frame under earthquake loading, are fixed or prescribed from outside, whereas a learning model changes its own parameters, and it does so through a second dynamical system, the training algorithm. Training is a closed loop in the sense of control theory: it adjusts the parameters until the output reaches a goal, much as one could tune the length of a pendulum by feedback until it resonates with an external drive.

A language model is therefore an iterated map with dynamics in a high-dimensional space. Many of its phenomena can be described, explained and predicted with the terms, tools and methods of nonlinear dynamics, and some of its problems may be solved with them. This is an opportunity, because a large community has developed nonlinear dynamics for more than a century, and its body of knowledge can serve the young field of language-model research.

The two communities, however, are largely separate. Most researchers build their careers within one of the fields, and each field has its own literature, jargon, catalog of phenomena and conferences, so joint work is rare.

The cost of this separation is that well-understood dynamics is rediscovered under new names. Nonlinear dynamics has a mature theory for the behaviors that appear in language models, built on pendulums, rotating machinery, ships, lasers and networks of coupled oscillators: stability and bifurcations, synchronization, chaos and Lyapunov exponents, and noise-driven escape from potential wells \citep{strogatz2018nonlinear,kramers1940brownian,hanggi1990reaction,acebron2005kuramoto}. These tools transfer directly. Escape from a repetition loop at sampling temperature $T$ is the escape of a particle from a potential well under noise of intensity $T$, with the gap between the logits of the repeated token and its best competitor in the role of the barrier. The Arrhenius form of the Kramers rate then predicts how the escape probability depends on temperature, and the measured growth of the barrier predicts how long a loop survives (Section~\ref{sec:loops} and Experiment~3). The exchange also runs the other way. Neural networks already forecast spatiotemporally chaotic systems from data \citep{pathak2018model} and learn coordinates in which nonlinear dynamics becomes linear \citep{lusch2017deep}, and a trained transformer is itself a high-dimensional, non-autonomous system that can be run millions of times, observed completely and differentiated exactly.

One goal of this paper is therefore to build a bridge between the two communities and to show where collaboration and joint research can pay off. For this reason we define the terms and methods of both fields that the argument needs, including some basic ones, so that members of each community can follow the other side. Appendix~\ref{app:glossary} collects these definitions in one place, and Appendix~\ref{app:notation} lists the notation.

A bridge between two fields can carry three kinds of traffic. The first is description, in which a behavior seen in one field receives a model, a name and a set of measurable quantities from the other, so that a repetition loop becomes a periodic orbit and a step-to-step oscillation of the loss becomes a period-2 orbit of a map. The second is explanation, in which the tools of the older field account for why the behavior occurs and predict how it changes with a parameter, as Kramers' theory predicts how the rate of escape from a well changes with the noise. The third is solution, in which the explanation points to an intervention that fixes a problem the newer field had not solved. Most of this paper works at the first two levels.

\subsection{A language model as three coupled dynamical systems}
\label{sec:threemodes}

A language model is used as an iterated map in three modes:
\begin{enumerate}[leftmargin=*,itemsep=1pt,topsep=2pt]
\item the forward pass, which carries the representation of a token through the layers $\ell$;
\item training, which updates the weights one optimizer step $t$ at a time;
\item generation, which chains forward passes to generate text one token $n$ at a time.
\end{enumerate}

The forward pass iterates over the depth of the model. Each residual block adds an update to the hidden state of every token, so a stack of $L$ blocks is a discrete-time dynamical system with the layer index as time \citep{e2017proposal,haber2017stable,lu2017finite,chen2018neural}, and attention couples the tokens into a system of interacting particles \citep{lu2019improving,sander2021sinkformers,geshkovski2023mathematical}. Attention acts like a force field: it attracts and repels the representations of the tokens as they flow through the layers of the model.

Training iterates over optimizer steps. The state is the parameter vector $\theta$, and gradient descent, $\theta_{t+1}=\theta_t-\eta\nabla\mathcal L(\theta_t)$, is a map on parameter space. It is the explicit Euler discretization of the gradient flow $\dot\theta=-\nabla\mathcal L(\theta)$, which describes an overdamped marble sliding down the loss landscape toward a minimum. Like any discretized flow, this map has dynamics of its own: at the step sizes used in practice it sits at the boundary of linear stability \citep{cohen2021gradient}, it produces loss spikes \citep{chowdhery2022palm,wortsman2023smallscale}, and it goes through abrupt transitions such as grokking \citep{power2022grokking} and the formation of induction heads \citep{olsson2022incontext}.

Generation iterates over the generated tokens. An autoregressive model feeds its own samples back as input, so decoding is a stochastic dynamical system on the space of contexts, and its best known pathology, the repetition loop \citep{holtzman2019curious,xu2022learning}, has recently been analyzed as an attractor \citep{wang2025unveiling,jones2026chaos}.

Each mode advances its own time index. We therefore call the modes the \emph{three clocks} of a language model: clock~1 is the forward pass, with layer time $\ell$; clock~2 is training, with training time $t$; and clock~3 is generation, with generation time $n$ (Figure~\ref{fig:clocks}). The clocks are coupled, since the parameters produced by training define the map that the forward pass iterates, and every tick of generation calls a full forward pass. They also run at very different rates, since the weights do not move during generation and each weight update averages over many forward passes. Separated time scales of this kind are the standard setting of slow-fast dynamical systems \citep{strogatz2018nonlinear}, and they help explain why three literatures grew up around one object with few citations between them.

\subsection{This work}
\label{sec:thiswork}

In the current work we build this bridge in both directions. We write the three clocks in the language of nonlinear dynamics, record which correspondences are proved, measured or only conjectured, and use the tools of the field to explain behavior that the machine learning literature documents but does not yet explain.

At the first level, we show each correspondence directly: each concept is drawn in a mechanical system and in a language model in the same representation, such as the same time history, spectrum or escape plot. At the second level, reduced-order models explain each correspondence, in the way an engineer replaces a sloshing tank or a rolling ship by an equivalent oscillator. The softmax turns out to be a Boltzmann distribution, gradient descent an Euler step on a nonlinear spring, a loss spike a ball thrown out of a stiff valley, attention on the circle a Kuramoto model, and a repetition loop a well that deepens while it is occupied. At the third level, we test on GPT-2 three controllers derived from the escape analysis. They remove exact loops and find the temperature window without a sweep, but none produces more coherent text than the best fixed temperature, because the trap is stored in the context while the controllers act on the noise (\op{12}). Solutions at this level remain the main open task, and the open problems of Section~\ref{sec:open} list the places where we expect them. Our contributions are the following.
\begin{enumerate}[leftmargin=*,itemsep=2pt]
\item \textbf{A unified framework} that writes the three clocks as coupled maps and organizes about 200 papers from machine learning theory, interpretability, optimization, statistical physics and applied mathematics around shared dynamical concepts, each shown in a physical system and a language model side by side.
\item \textbf{The Three-Clock Dictionary} (Table~\ref{tab:dictionary}), which maps fifteen concepts from nonlinear dynamics onto each clock with the evidence level of every entry; each empty cell defines a research question.
\item \textbf{Reduced-order models and theory}, including an exact Arrhenius law for escape from repetition loops and conditions under which a loop locks.
\item \textbf{Experiments on open models} that run on a laptop CPU, one per clock, and \textbf{twelve open problems}, each stated with a first test.
\end{enumerate}

\paragraph{Organization.} Section~\ref{sec:maps} defines the three clocks as maps and reads each one as a dynamical system. Section~\ref{sec:dictionary} opens with the Three-Clock Dictionary and then goes through its main concepts, from stability and basins to synchronization, noise-driven escape and chaos, each shown in a mechanical system and in a language model side by side. Section~\ref{sec:reduced} explains why the pairs behave alike by reducing each behavior of the language model to a physical model with the same governing term, and proves when a repetition loop locks. Section~\ref{sec:experiments} tests three of these correspondences in new experiments, one per clock: whether training changes how the forward pass stretches perturbations, how gradient descent on a transformer moves from a period-2 orbit to irregular oscillation as the step size grows, and whether escape from repetition loops follows the Arrhenius law with a barrier that deepens over time. Section~\ref{sec:open} states twelve open problems, and the appendices give the glossary, the notation, the experimental details and further work on each clock.

\paragraph{Related work and scope.} Several papers develop one of these views in depth, and we rely on them throughout: attention as an interacting particle system over depth \citep{geshkovski2023mathematical,castin2025unified}, gradient descent at the edge of stability \citep{cohen2021gradient}, and stochastic and chaotic accounts of generation \citep{krishnamurthy2026stochastic,jones2026chaos}. To our knowledge, no earlier work treats the three clocks side by side in one vocabulary, with an explicit record of which correspondences are proved, which are measured and which are conjectures. We use the language of dynamical systems in its technical sense wherever a result supports it, and we flag analogies as analogies; for example, we call a loss spike a bifurcation only when a parameter can be shown to cross a stability threshold.

\begin{tcolorbox}[breakable, enhanced jigsaw, colback=white, colframe=black, boxrule=0.5pt, arc=0pt, left=5pt, right=5pt, top=3pt, bottom=3pt]
\textbf{Practical takeaways.}
\begin{itemize}[leftmargin=*,itemsep=1pt,topsep=2pt]
\item \emph{Generation.} In GPT-2 and Pythia, coherent text appears only in a narrow band of temperatures between repetition and incoherence, and for some prompts in none. A loop is not locked by a fixed temperature: the temperature needed to leave it within a given budget falls like $1/\ln N$ with the length $N$ of the generation, and larger models need a higher temperature to leave the same kind of loop (Section~\ref{sec:exp3}). In Qwen3-0.6B on GSM8K, accuracy within a fixed budget peaks at the vendor's $T=0.6$, and below it the model rethinks the same step until the budget runs out.
\item \emph{Anti-repetition.} Raising the temperature when a loop starts, rolling the text back to the start of the loop, or steering the temperature by feedback all remove exact loops, but none beats the best fixed temperature on coherent text, because the attractor is stored in the context, not in the sampler (\op{12}).
\item \emph{Training.} At the edge of stability several Hessian eigenvalues sit 5\% to 14\% above $2/\eta$ while the curvature along the actual step equals $2/\eta$, so $\eta\lambda_{\max}>2$ alone does not signal divergence. A power iteration can underestimate $\lambda_{\max}$ there, because the top eigenvalues form a cluster that reorders at every step (Section~\ref{sec:exp2}).
\item \emph{Depth.} Training makes each layer stretch perturbations more, while the stretched directions of consecutive layers become less aligned (Section~\ref{sec:exp1}).
\item \emph{Tools.} Per-layer Lyapunov exponents, learning-rate bifurcation diagrams and exact loop-escape probabilities from a single greedy pass all run on a laptop CPU, and the code is released.
\end{itemize}
\end{tcolorbox}

\section{The three clocks}
\label{sec:maps}
\label{sec:primer}

Section~\ref{sec:intro} described three ways in which a language model is iterated. This section writes them as maps, so that each can be read as a dynamical system with a state, a time and a rule of evolution. It then explains how the three maps are nested and why they run at different speeds, and it ends with the places where the dynamical reading needs care. The rest of the paper uses these definitions throughout.

Consider a decoder-only transformer \citep{vaswani2017attention} with $L$ residual blocks, model width $d$ and vocabulary $\mathcal{V}$. For a context of $n$ tokens, write $x^{\ell}=(x^{\ell}_1,\dots,x^{\ell}_n)\in\mathbb{R}^{n\times d}$ for the residual stream after block $\ell$, and $\theta$ for all parameters. The model defines three maps. Appendix~\ref{app:notation} lists all symbols used in the paper.

\begin{definition}[Clock 1: layer time]
\label{def:clock1}
For fixed $\theta$ and fixed input tokens,
\begin{equation}
x^{\ell+1} = x^{\ell} + f^{\ell}_{\theta}(x^{\ell}), \qquad \ell = 0,\dots,L-1,
\label{eq:clock1}
\end{equation}
where $f^{\ell}_\theta$ contains attention, which couples token $i$ to tokens $j\le i$, and a token-wise MLP. The logits are $z = W_U\, \mathrm{LN}(x^{L}_n)$.
\end{definition}

\paragraph{Clock 1 read as a dynamical system.} Equation~\eqref{eq:clock1} is a forward Euler step, with unit step size, of an ordinary differential equation $\dot x = f(x,\ell)$ for $n$ particles in $\mathbb{R}^d$. The tokens are the particles, their hidden states are the positions, and attention is the force through which they interact. The system is non-autonomous, since each block has its own weights, and it runs for a fixed and small number of steps. This observation was made early for residual networks \citep{he2015deep} by \citet{e2017proposal}, \citet{haber2017stable} and \citet{lu2017finite}, and it led to continuous-depth models \citep{chen2018neural}, to architectures designed from the stability theory of ODEs and PDEs \citep{haber2017stable,ruthotto2018deep,chang2019antisymmetricrnn}, to invertible second-order variants \citep{sander2021momentum}, and to a view of training as optimal control \citep{e2018meanfield}. Whether a trained residual network actually converges to an ODE as depth grows depends on the initialization and on how the residual branch is scaled \citep{sander2022residual}. In transformers, several stabilization methods amount to a choice of this step size, either by scaling the residual branch at initialization \citep{bachlechner2020rezero,wang2022deepnet} or by moving the layer normalization inside the branch \citep{xiong2020layer,liu2020difficulty}.

\begin{definition}[Clock 2: training time]
\label{def:clock2}
For a loss $\mathcal{L}$ and a stream of minibatches $\mathcal{B}_t$,
\begin{equation}
\theta_{t+1} = \theta_t - \eta_t\, P_t\, \nabla_\theta \mathcal{L}(\theta_t;\mathcal{B}_t), \qquad t = 0,1,\dots
\label{eq:clock2}
\end{equation}
where $P_t$ is a preconditioner (identity for gradient descent, a diagonal matrix built from gradient moments for Adam) and $\mathcal{L}(\cdot;\mathcal{B}_t)$ is the loss on minibatch $\mathcal{B}_t$. With full batches the map is deterministic; with minibatches it is a random dynamical system.
\end{definition}

\paragraph{Clock 2 read as a dynamical system.} With $P_t$ the identity, \eqref{eq:clock2} is the explicit Euler scheme for the gradient flow $\dot\theta=-\nabla\mathcal L(\theta)$, the motion of an overdamped particle in the potential $\mathcal L$, with the learning rate $\eta$ as time step. The state is the parameter vector, with millions to billions of entries. The data stream is the sequence of minibatches presented to the optimizer, one per step. Because each minibatch is drawn at random, the map is driven by noise, and because the data, the schedule $\eta_t$ and the preconditioner change over training, it is non-autonomous.

\begin{definition}[Clock 3: generation time]
\label{def:clock3}
For fixed $\theta$, a sampling temperature $T>0$ and a context $s_n=(y_1,\dots,y_n)$,
\begin{equation}
y_{n+1}\sim \pi_T(\cdot\mid s_n) = \mathrm{softmax}\big(z(s_n;\theta)/T\big), \qquad s_{n+1} = (s_n, y_{n+1}).
\label{eq:clock3}
\end{equation}
This is a Markov chain on the growing space of contexts. Truncation rules such as top-k or nucleus sampling change $\pi_T$ but not the structure.
\end{definition}

\paragraph{Clock 3 read as a dynamical system.} Each step chooses one of $|\mathcal V|$ tokens, and the chosen token is appended to the input of the next step, so the output is fed back as input. Definition~\ref{def:clock3} makes generation a Markov chain whose state is the whole context. The state space grows with $n$, so the usual results on stationary distributions do not apply directly, but the questions of dynamical systems still make sense: which sets of contexts trap the chain, how long the chain stays in them, and how these answers depend on the temperature $T$ and on truncation. \citet{krishnamurthy2026stochastic} formulate language models as high-dimensional nonlinear autoregressive models and analyze networks of interacting models with mean-field equations. A deterministic version of clock~3, obtained by feeding back the expected embedding of the next-token distribution instead of a sample, is a continuous map that can be analyzed with standard tools \citep{jones2026chaos}. Temperature divides the logits by $T$, so $\pi_T$ in \eqref{eq:clock3} is a Gibbs distribution with energy $-z$ at temperature $T$. Greedy decoding is the limit $T\to0$, and $T$ sets how often the chain leaves the most likely path, so it serves as both the noise intensity and the natural bifurcation parameter of clock~3. Its effect on task accuracy \citep{renze2024effect} and on novelty and coherence \citep{peeperkorn2024temperature} has been measured directly.

\paragraph{Nested clocks and separated time scales.} The three maps are nested. Evaluating $z(s_n;\theta)$ in \eqref{eq:clock3} means running \eqref{eq:clock1} to the end, and computing the gradient in \eqref{eq:clock2} means running \eqref{eq:clock1} on teacher-forced contexts, taken from the training text and not from the model's own samples. The weights produced by \eqref{eq:clock2} define the maps in \eqref{eq:clock1} and \eqref{eq:clock3}. The clocks also run at very different speeds. Generation never changes $\theta$, and one training step changes it by an amount proportional to the small learning rate $\eta$, so the forward map changes only by $O(\eta)$ per step, an amount bounded by a constant times $\eta$. Counted in optimizer steps $t$, the weights therefore evolve on the slow time $\eta t$, while each forward pass and each generation runs to completion at frozen weights. In the notation of the method of multiple scales, with $\varepsilon=\eta$, the fast time is $\tau_0=t$ and the weights depend on the slow time $\tau_1=\varepsilon t$, and \citet{berthier2026multiple} carry this expansion to three time scales, $\tau_2=\varepsilon^2t$ included, for gradient descent at the edge of stability. In the language of slow-fast systems, clock~2 is the slow variable and clocks~1 and~3 are fast dynamics evaluated at a frozen slow state \citep{strogatz2018nonlinear}.

Two well-studied effects break this separation. With in-context learning the fast clock imitates the slow one: for in-context regression, a linear self-attention layer can implement one step of gradient descent on examples given in the prompt, so a stack of layers runs an optimizer whose iterations are the layers \citep{oswald2022transformers,akyurek2022learning,garg2022transformers}. Trained linear transformers learn preconditioned versions of this algorithm \citep{ahn2023transformers}, and the gradient-flow dynamics that produce them can be solved in closed form \citep{zhang2023trained}. In self-training the fast clock drives the slow one: text produced by clock~3 is used as training data for clock~2 \citep{shumailov2023curse,bertrand2023stability}, which we discuss in Appendix~\ref{app:related}.

\paragraph{Where the analogies are weak.} First, clock~1 runs for a fixed and small number of steps (between 6 and about 100 layers), so asymptotic notions such as attractors and Lyapunov exponents must be replaced by finite-time versions. Second, clock~2 is driven by a data stream whose statistics are not stationary in general, and by learning-rate schedules, which makes it a non-autonomous system. Third, the state of clock~3 is the whole context, so its dimension grows with time and the chain never returns to an earlier state. Statements about loops and attractors on clock~3 are therefore about the projection of the state onto recent tokens. Each of these differences limits which theorems can be imported from the classical theory without change.

\section{Shared behavior: a dictionary between the clocks and nonlinear dynamics}
\label{sec:dictionary}
\label{sec:tools}

With the clocks written as maps, the next question is which behaviors of nonlinear systems they show. Table~\ref{tab:dictionary} answers it for fifteen concepts at once, and it serves as the map of this section. Each row is a concept from nonlinear dynamics and each column is a clock. A cell names the observed behavior that the concept describes on that clock and marks its evidence level: \statusP\ when the correspondence is proved for a model of the system, \statusM\ when it has been measured in trained networks, and \statusC\ when it is a conjecture. A question mark means that we found no work that tests the concept on that clock.

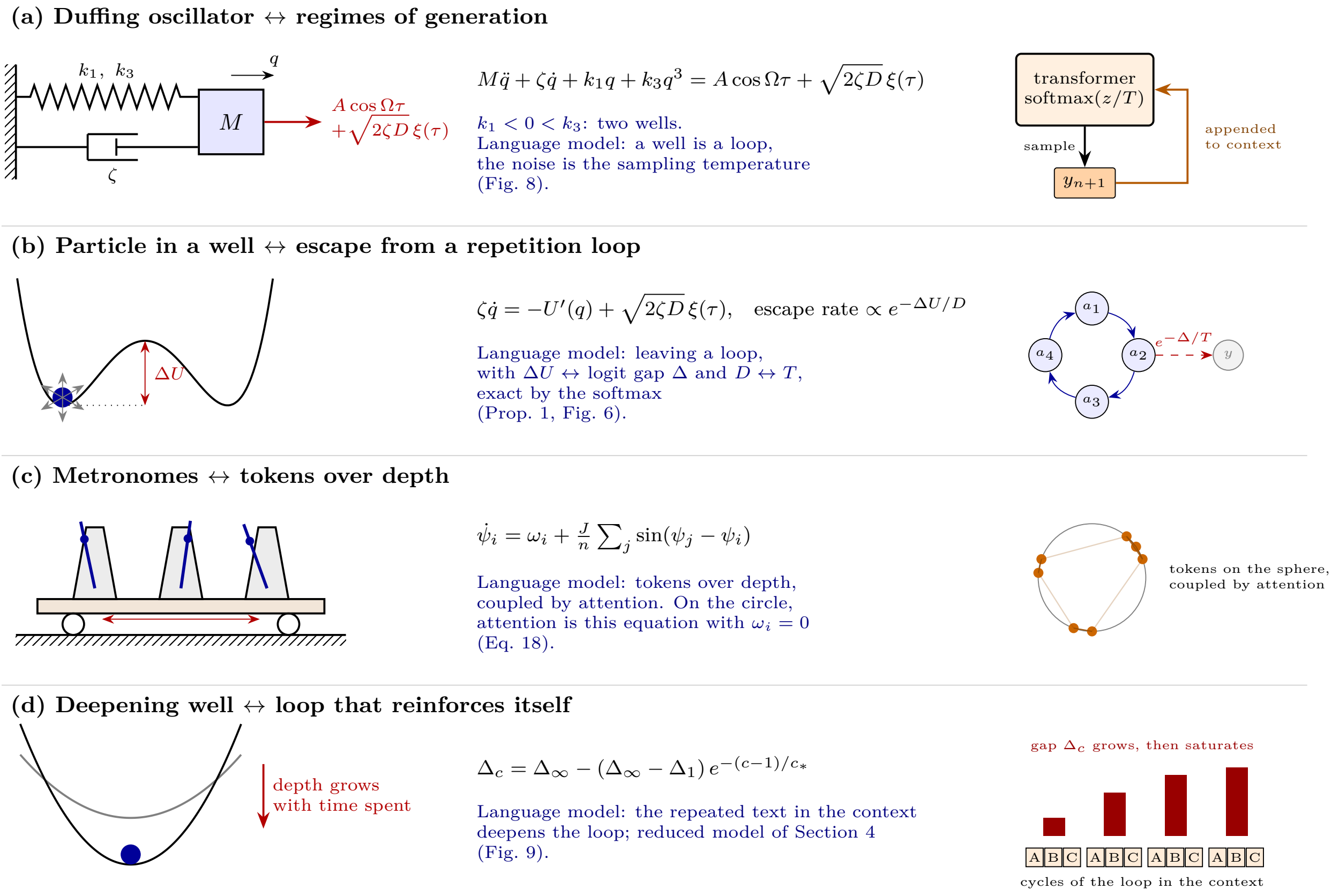
\begin{figure}[!t]
\centering
\resizebox{\textwidth}{!}{%
\begin{tikzpicture}[>=Stealth, font=\small,
  mass/.style={draw, thick, fill=blue!10, minimum width=9mm, minimum height=9mm},
  spring/.style={decorate, decoration={zigzag, amplitude=1.6mm, segment length=2.2mm, pre length=2mm, post length=2mm}, thick},
  ground/.style={pattern=north east lines, draw=none}]
\def\eqx{6.3}
\begin{scope}[shift={(0,0)}]
\node[anchor=west, font=\small\bfseries] at (-0.2,2.45) {(a) Duffing oscillator $\leftrightarrow$ regimes of generation};
\fill[ground] (-0.15,0.2) rectangle (0,1.8); \draw[thick] (0,0.2) -- (0,1.8);
\node[mass] (m) at (3.0,1.0) {$M$};
\draw[spring] (0,1.35) -- node[above=1.2mm, font=\scriptsize] {$k_1,\ k_3$} ($(m.west)+(0,0.35)$);
\draw[thick] (0,0.65) -- (1.0,0.65); \draw[thick] (1.0,0.47) -- (1.0,0.83); \draw[thick] (1.0,0.47) -- (1.7,0.47) (1.0,0.83) -- (1.7,0.83);
\draw[thick] (1.4,0.53) -- (1.4,0.77); \draw[thick] (1.4,0.65) -- ($(m.west)+(0,-0.35)$); \node[font=\scriptsize] at (1.35,0.25) {$\zeta$};
\draw[->, thick, red!70!black] (m.east) -- ++(0.8,0) node[right, font=\scriptsize, align=left] {$A\cos\Omega\tau$\\$+\sqrt{2\zeta D}\,\xi(\tau)$};
\draw[->] (3.0,1.65) -- ++(0.6,0) node[above, font=\scriptsize] {$q$};
\node[anchor=west, font=\footnotesize] at (\eqx,1.6) {$M\ddot q+\zeta\dot q+k_1q+k_3q^3=A\cos\Omega\tau+\sqrt{2\zeta D}\,\xi(\tau)$};
\node[anchor=north west, font=\scriptsize, text=blue!50!black, align=left] at (\eqx,1.2) {$k_1<0<k_3$: two wells.\\Language model: a well is a loop,\\the noise is the sampling temperature\\(Fig.~\ref{fig:atlas}).};
\begin{scope}[shift={(14.0,0)}]
\node[draw, thick, fill=orange!12, rounded corners=2pt, minimum width=1.9cm, minimum height=1.0cm, font=\scriptsize, align=center] (tf) at (0.9,1.45) {transformer\\$\mathrm{softmax}(z/T)$};
\node[draw, fill=orange!35, rounded corners=1pt, font=\scriptsize, minimum height=0.42cm] (yn) at (0.9,0.15) {$y_{n+1}$};
\draw[->, thick] (tf.south) -- node[left, font=\tiny] {sample} (yn.north);
\draw[->, thick, orange!70!black] (yn.east) -- ++(1.0,0) |- (tf.east);
\node[font=\tiny, align=left, anchor=west, text=orange!50!black] at (2.45,0.8) {appended\\to context};
\end{scope}
\draw[gray!35] (-0.2,-0.45) -- (18.0,-0.45);
\end{scope}
\begin{scope}[shift={(0,-3.2)}]
\node[anchor=west, font=\small\bfseries] at (-0.2,2.45) {(b) Particle in a well $\leftrightarrow$ escape from a repetition loop};
\draw[thick, domain=-1.55:1.55, samples=60, smooth] plot ({1.15*\x+1.8}, {0.9*(\x*\x-1)*(\x*\x-1)+0.25});
\fill[blue!60!black] (0.65,0.36) circle (1.4mm);
\draw[<->, red!70!black] (1.8,0.25) -- (1.8,1.15) node[midway, right, font=\scriptsize] {$\Delta U$};
\draw[dotted] (0.65,0.25) -- (1.8,0.25);
\foreach \a in {30,90,150,210,270,330}{\draw[->, gray, thin] (0.65,0.36) -- ++(\a:0.35);}
\node[anchor=west, font=\footnotesize] at (\eqx,1.6) {$\zeta\dot q=-U'(q)+\sqrt{2\zeta D}\,\xi(\tau)$,\quad escape rate $\propto e^{-\Delta U/D}$};
\node[anchor=north west, font=\scriptsize, text=blue!50!black, align=left] at (\eqx,1.2) {Language model: leaving a loop,\\with $\Delta U\leftrightarrow$ logit gap $\Delta$ and $D\leftrightarrow T$,\\exact by the softmax\\(Prop.~\ref{prop:arrhenius}, Fig.~\ref{fig:pair_escape}).};
\begin{scope}[shift={(15.0,0.95)}]
\foreach \i/\ang in {1/90,2/0,3/270,4/180}{\node[circle, draw, fill=blue!8, inner sep=0pt, minimum size=4.6mm, font=\tiny] (b\i) at (\ang:0.65) {$a_{\i}$};}
\foreach \i/\j in {1/2,2/3,3/4,4/1}{\draw[->, blue!60!black] (b\i) to[bend left=30] (b\j);}
\node[circle, draw=gray, fill=gray!10, inner sep=0pt, minimum size=4.2mm, font=\tiny, text=gray] (ex) at (1.9,0.0) {$y$};
\draw[->, dashed, red!70!black] (b2) -- node[above, font=\tiny, text=red!70!black] {$e^{-\Delta/T}$} (ex);
\end{scope}
\draw[gray!35] (-0.2,-0.45) -- (18.0,-0.45);
\end{scope}
\begin{scope}[shift={(0,-6.4)}]
\node[anchor=west, font=\small\bfseries] at (-0.2,2.45) {(c) Metronomes $\leftrightarrow$ tokens over depth};
\draw[thick, fill=brown!20] (0.3,0.55) rectangle (4.3,0.75);
\foreach \x in {0.8,3.8}{\draw[thick] (\x,0.4) circle (0.15);}
\fill[ground] (0,0.1) rectangle (4.6,0.25); \draw[thick] (0,0.25) -- (4.6,0.25);
\foreach \x/\ang in {1.1/12,2.3/-8,3.5/20}{
  \draw[thick, fill=gray!15] (\x-0.3,0.75) -- (\x+0.3,0.75) -- (\x+0.12,1.75) -- (\x-0.12,1.75) -- cycle;
  \draw[very thick, blue!60!black] (\x,0.9) -- ++(90+\ang:0.95); \fill[blue!60!black] (\x,0.9) ++(90+\ang:0.7) circle (0.06);}
\draw[<->, red!70!black] (1.2,0.47) -- (3.4,0.47);
\node[anchor=west, font=\footnotesize] at (\eqx,1.6) {$\dot\psi_i=\omega_i+\frac{J}{n}\sum_j\sin(\psi_j-\psi_i)$};
\node[anchor=north west, font=\scriptsize, text=blue!50!black, align=left] at (\eqx,1.2) {Language model: tokens over depth,\\coupled by attention. On the circle,\\attention is this equation with $\omega_i=0$\\(Eq.~\ref{eq:attn_kuramoto}).};
\begin{scope}[shift={(15.0,1.05)}]
\draw[gray, thin] (0,0) circle (0.75);
\foreach \a in {20,35,50,160,175,250,270}{\fill[orange!80!black] (\a:0.75) circle (0.07);}
\foreach \a/\b/\w in {20/35/0.9,20/50/0.8,35/50/0.9,160/175/0.9,250/270/0.9,50/160/0.25,175/250/0.25,20/270/0.25}{\draw[orange!60!black, opacity=\w, line width=0.5pt] (\a:0.75) -- (\b:0.75);}
\node[font=\tiny, align=left, anchor=west] at (0.95,0) {tokens on the sphere,\\coupled by attention};
\end{scope}
\draw[gray!35] (-0.2,-0.45) -- (18.0,-0.45);
\end{scope}
\begin{scope}[shift={(0,-9.6)}]
\node[anchor=west, font=\small\bfseries] at (-0.2,2.45) {(d) Deepening well $\leftrightarrow$ loop that reinforces itself};
\draw[thick, gray, domain=-1.4:1.4, samples=50, smooth] plot ({1.1*\x+1.6}, {0.45*(\x*\x)+0.9});
\draw[thick, domain=-1.4:1.4, samples=50, smooth] plot ({1.1*\x+1.6}, {1.0*(\x*\x)+0.25});
\fill[blue!60!black] (1.6,0.39) circle (1.4mm);
\draw[->, red!70!black, thick] (3.45,1.65) -- (3.45,0.75) node[midway, right, font=\scriptsize, align=left] {depth grows\\with time spent};
\node[anchor=west, font=\footnotesize] at (\eqx,1.6) {$\Delta_c=\Delta_\infty-(\Delta_\infty-\Delta_1)\,e^{-(c-1)/c_\ast}$};
\node[anchor=north west, font=\scriptsize, text=blue!50!black, align=left] at (\eqx,1.2) {Language model: the repeated text in the context\\deepens the loop; reduced model of Section~\ref{sec:reduced}\\(Fig.~\ref{fig:reduced}).};
\begin{scope}[shift={(14.2,0.35)}]
\foreach \c/\h in {0/0.25,1/0.6,2/0.85,3/0.95}{
  \foreach \k in {0,1,2}{\node[draw, fill=orange!15, minimum width=0.24cm, minimum height=0.26cm, inner sep=0pt, font=\tiny] at (0.85*\c+0.26*\k,0) {\pgfmathparse{{"A","B","C"}[\k]}\pgfmathresult};}
  \fill[red!60!black] (0.85*\c+0.12,0.3) rectangle (0.85*\c+0.42,0.3+\h);}
\node[font=\tiny, text=red!60!black] at (1.5,1.55) {gap $\Delta_c$ grows, then saturates};
\node[font=\tiny] at (1.5,-0.35) {cycles of the loop in the context};
\end{scope}
\end{scope}
\end{tikzpicture}}
\caption{Four physical systems and their language-model twins. In each row the mechanical system (left) and the language model (right) look nothing alike, yet they share the term of the governing equation (middle) that produces the behavior; Section~\ref{sec:reduced} derives each correspondence. (a) A mass $M$ on a spring with linear stiffness $k_1$ and cubic stiffness $k_3$, friction coefficient $\zeta$, harmonic forcing of amplitude $A$ and frequency $\Omega$, and white noise $\xi$ of intensity $D$; with $k_1<0$ the potential $\tfrac12k_1q^2+\tfrac14k_3q^4$ has two wells. (b) An overdamped particle in a potential well under thermal noise of intensity $D$, whose escape rate over the barrier $\Delta U$ is given by Kramers' formula \eqref{eq:kramers}. (c) Metronomes on a board that can roll, which synchronize through the small motion of the board; their phases follow the Kuramoto model \eqref{eq:kuramoto}. (d) A well whose depth grows with the time the particle has spent in it and then saturates.}
\label{fig:models}
\end{figure}

\begin{table}[p]
\centering
\scriptsize
\setlength{\tabcolsep}{3pt}
\renewcommand{\arraystretch}{1.25}
\begin{tabularx}{\textwidth}{>{\raggedright\arraybackslash}p{2.1cm} >{\raggedright\arraybackslash}X >{\raggedright\arraybackslash}X >{\raggedright\arraybackslash}X}
\toprule
\textbf{Concept} & \textbf{Clock 1: layer time $\ell$} & \textbf{Clock 2: training time $t$} & \textbf{Clock 3: generation time $n$} \\
\midrule
Fixed point & Tokens collapse to a cluster; rank collapse \statusP\ \citep{geshkovski2023mathematical,dong2021attention}. Massive activations nearly input independent \statusM\ \citep{sun2024massive} & Convergence to a minimum; neural collapse at the end of training \statusM\ \citep{papyan2020prevalence} & Period-one loops (one token repeated) as the simplest case of degenerate repetition \statusM\ \citep{holtzman2019curious} \\
Linear stability, Jacobian spectrum & Dynamical isometry; layer Jacobians of trained models \statusP/\statusM\ \citep{pennington2017resurrecting,fernando2026dynamics} & Sharpness vs.\ $2/\eta$ \statusM\ \citep{cohen2021gradient} & Derivative of the feedback map factorizes through $1/T$ \statusM\ \citep{jones2026chaos} \\
Finite-time Lyapunov exponent & Ridges in input space separate classes \statusM\ \citep{storm2023finitetime}; per-layer profile over training \statusM\ (Sec.~\ref{sec:exp1}) & Sensitivity of runs to minibatch noise early in training \statusM\ \citep{frankle2019linear} & Divergence of nearby generations \statusM\ \citep{jaca2026chaos,jones2026chaos} \\
Edge of chaos, criticality & Order-chaos transition at initialization sets trainable depth \statusP\ \citep{poole2016exponential,schoenholz2016deep} & Edge of stability \statusM\ \citep{cohen2021gradient} & Critical temperature with power-law text statistics \statusM\ \citep{nakaishi2024phase} \\
Limit cycle (periodic orbit) & \na & Period-2 orbits of GD \statusP\ \citep{chen2022edge}; SGD converges to cycles \statusP\ \citep{chaudhari2017stochastic} & Repetition loops \statusM\ \citep{xu2022learning}; paraphrase 2-cycles \statusM\ \citep{wang2025unveiling}; loops converge in hidden-state space \statusM\ (Exp.~3) \\
Period doubling, route to chaos & \na & Larger step sizes give period doubling and chaos in minimal models \statusP\ \citep{zhu2022edgeofstability,kong2020stochasticity}; period-2 orbit, then irregular oscillation, in a transformer \statusM\ (Exp.~2) & Feigenbaum scaling in embedding feedback \statusM\ \citep{jones2026chaos} \\
Fractal boundary & \na & Trainable vs.\ divergent hyperparameters \statusM\ \citep{sohldickstein2024boundary}; interleaved trainable and divergent step sizes \statusM\ (Exp.~2); fractal attractors of optimizers \statusM\ \citep{tuci2026generalization} & \na \\
Basin of attraction & Class regions bounded by FTLE ridges \statusM\ \citep{storm2023finitetime} & Which minimum is reached depends on step size and batch size \statusM\ \citep{jastrzebski2017three} & Prompts that do or do not fall into loops \statusC \\
Metastability, slow manifold & Multi-cluster states last exponentially long \statusP\ \citep{geshkovski2024dynamic} & Plateaus \statusP\ \citep{saxe2013exact}; slow phase of pretraining \statusM\ \citep{zhang2026stability,wen2024warmupstabledecay} & Long coherent stretches before drift \statusC \\
Saddle and ghost & \na & Saddles dominate \statusM\ \citep{dauphin2014identifying}; abrupt learning from a ghost \statusP\ \citep{dinc2025ghost} & \na \\
Phase transition & \na & Grokking \statusP/\statusM\ \citep{rubin2023grokking,power2022grokking}; induction heads \statusM\ \citep{olsson2022incontext} & Low-$T$ repetitive vs.\ high-$T$ incoherent phase \statusM\ \citep{nakaishi2024phase,arnold2024phase}; narrow coherent window between them, also in a reasoning model \statusM\ (Exp.~3) \\
Noise-driven (Kramers) escape & \na & Escape of SGD from minima \statusP\ \citep{xie2020diffusion,mori2021powerlaw} & Escape from repetition loops, logit gap as barrier \statusP/\statusM\ (Prop.~\ref{prop:arrhenius}, Exp.~3); locking temperature \statusP\ (Prop.~\ref{prop:tlock}), horizon-dependent in practice \statusM\ (Exp.~3) \\
Synchronization, consensus & Clustering as consensus; Kuramoto form \statusP\ \citep{geshkovski2023mathematical,li2026diverse}; synchronization by noise \statusP\ \citep{agazzi2026stochastic} & \na & Collective dynamics of interacting models \statusP\ \citep{krishnamurthy2026stochastic} \\
Koopman, linearization & Local linear description of each layer \statusM\ \citep{fernando2026dynamics} & Koopman-based training and pruning \statusM\ \citep{dogra2020optimizing,redman2021operator}; lazy regime \statusP\ \citep{jacot2018neural} & \na \\
Conservation law, symmetry & Gradient-flow structure of attention \statusP\ \citep{sander2021sinkformers} & Symmetries give conserved quantities that finite steps break \statusP\ \citep{kunin2020neural} & \na \\
\bottomrule
\end{tabularx}
\caption{The Three-Clock Dictionary. Each cell gives the behavior that a dynamical-systems concept describes on one clock, with its evidence level (\statusP: proved in a model; \statusM: measured in trained networks; \statusC: conjecture) and representative references. Question marks are cells for which we found no study; they define the open problems of Section~\ref{sec:open}.}
\label{tab:dictionary}
\end{table}

\paragraph{Reading the table.} Fixed points, stability and Lyapunov exponents have been used on all three clocks, although with different meanings: on clock~1 they describe a single finite trajectory, on clock~2 the long-run behavior of an optimizer, and on clock~3 a random process. Phase transitions and escape appear on clocks~2 and~3 but not on clock~1, where the number of steps is fixed and small, so there is no long-time limit in which a transition could be sharp. Even so, the metastability result of \citet{geshkovski2024dynamic} shows that finite-time transients on clock~1 can be analyzed with care. Clock~2 is the best covered column, because optimization theory has always been a theory of iterated maps. Clock~1 is covered mostly by results at initialization and by models with simplified attention. Clock~3 is the youngest column, and most of its entries date from 2024 or later.

The rest of this section goes through the rows that carry most of the evidence. Each subsection defines one concept, shows it in a mechanical system and in a language model side by side, and summarizes what is known on each clock. The order follows the way the concepts build on each other: the stability of a fixed point and its loss through period doubling (Section~\ref{sec:eos}), the basins that decide which state is reached and the slow transients near unstable states (Section~\ref{sec:transitions}), the synchronization of coupled units (Section~\ref{sec:particles}), escape from a well under noise (Section~\ref{sec:escape}), the ways to tell periodic, noisy and chaotic motion apart in a measured signal (Section~\ref{sec:regimes}), and the growth of small perturbations (Section~\ref{sec:edgechaos}). Figure~\ref{fig:models} shows the four physical systems that recur in this comparison next to their language-model counterparts. Section~\ref{sec:reduced} then explains why each pair behaves alike.

\subsection{Stability, period doubling and the edge of stability}
\label{sec:eos}

We start with the simplest question one can ask of an iterated map: does a fixed point attract nearby states, and what happens when it stops doing so?

\paragraph{Maps, fixed points and stability.} A discrete-time system is a map $u_{t+1}=F(u_t)$. A fixed point $u^\ast=F(u^\ast)$ is linearly stable if every eigenvalue of the Jacobian $\mathrm{D}F(u^\ast)$ lies inside the unit circle. The textbook case is the explicit Euler scheme for a stiff ordinary differential equation: integrating $\dot u=-\lambda u$ with step $\eta$ gives $u_{t+1}=(1-\eta\lambda)u_t$, which converges only if $\eta\lambda<2$ and otherwise oscillates with growing amplitude. Gradient descent with step size $\eta$ on a quadratic with Hessian eigenvalue $\lambda$ is the same map, and this one-line calculation is the origin of the threshold $2/\eta$ in the edge-of-stability literature discussed below.

\paragraph{Periodic orbits and bifurcations.} A period-$k$ orbit is a fixed point of $F^{k}$, and under a change of parameter such orbits can appear, disappear or exchange stability. In a period-doubling bifurcation an eigenvalue crosses $-1$ and a period-2 orbit is born. The logistic map $u_{t+1}=\mu u_t(1-u_t)$, introduced as a model of insect populations, passes through a cascade of such doublings to chaos as $\mu$ grows \citep{may1976simple}, and the ratio of successive bifurcation intervals tends to a universal constant, $4.669\ldots$, for every smooth one-humped map \citep{feigenbaum1978quantitative}. In a saddle-node bifurcation a pair of fixed points collides and disappears. Just past it the trajectory still slows down where the pair used to be (a ``ghost''), which produces long plateaus followed by fast jumps \citep{strogatz2018nonlinear,dinc2025ghost}. Gradient descent on a double-well loss $\mathcal{L}(u)=(u^2-1)^2/4$ already shows both behaviors as $\eta$ grows. The minimum loses stability at $\eta=1$, a doubling cascade follows, and near $\eta\approx1.6$ the chaotic iterate hops between the two wells.

\begin{figure}[t]
\centering
\includegraphics[width=\textwidth]{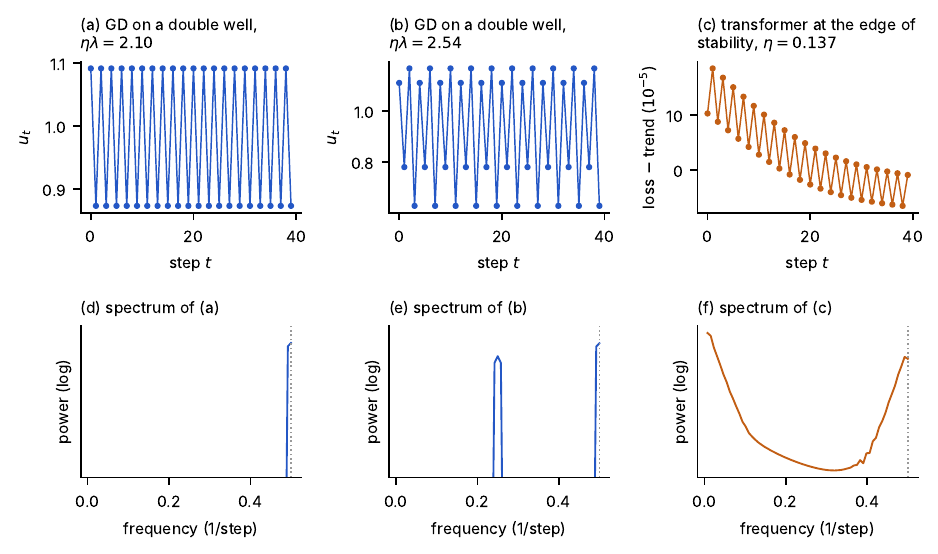}
\caption{Training as a dynamical system: period two at the edge of stability. (a, b) Gradient descent on the double well $\mathcal{L}(u)=(u^2-1)^2/4$, whose minimum has curvature $\lambda=2$, with $\eta\lambda=2.10$ and $2.54$: the last 40 iterates lie on a period-2 and a period-4 orbit. (d, e) Their power spectra, with peaks at frequencies $1/2$ and $1/4$. (c) The one-block transformer of Experiment~2 trained by full-batch gradient descent at $\eta=0.137$, at the edge of stability: training loss over the last 40 steps after removing a slow polynomial trend, in units of $10^{-5}$. The loss alternates from step to step. (f) Spectrum of the detrended loss over the last 400 steps: the same peak at $1/2$, on top of a low-frequency band from the slow modulation of the oscillation (Section~\ref{sec:exp2}).}
\label{fig:pair_training}
\end{figure}

\paragraph{On the clocks.} Classical optimization theory treats gradient descent as a discretization of gradient flow, which is valid when the step size is small compared with the inverse curvature. \citet{cohen2021gradient} showed that neural network training does not stay in that regime. Under full-batch gradient descent the sharpness, meaning the largest Hessian eigenvalue $\lambda_{\max}$, first rises (progressive sharpening) until it reaches $2/\eta$, and then hovers near that value while the loss keeps decreasing, though not monotonically. By the linear stability argument above, the iteration then sits on the boundary where the top direction loses stability. Adaptive methods show the same behavior with respect to the preconditioned sharpness \citep{cohen2022adaptive}.

For the regime as a whole there is now a continuous-time description. The central flow of \citet{cohen2024central}, an ODE for the time-averaged trajectory, keeps the sharpness at $2/\eta$ and predicts the loss curve of real networks, and \citet{berthier2026multiple} derive it by the method of multiple scales with three time scales. \citet{litman2026origin} shows that a weighted average of the curvature along the steps is forced to $2/\eta$. Bifurcation theory gives the local picture, with the onset of oscillation treated as a flip bifurcation near a manifold of minimizers (Appendix~\ref{app:related}). Section~\ref{sec:exp2} adds a measurement: in a small transformer the edge is multi-mode, with several eigenvalues above $2/\eta$, while the curvature along the step equals $2/\eta$.

At larger step sizes the map can become chaotic. Deterministic gradient descent on a multiscale objective, with a step size that resolves only the macroscopic scale, converges to a statistical distribution \citep{kong2020stochasticity}, and stochastic optimizers at the edge of stability have fractal attractors whose dimension has been related to generalization \citep{tuci2026generalization}. On clock~1 the corresponding question is whether a forward pass keeps signals at a constant size, which Section~\ref{sec:edgechaos} takes up, and on clock~3 it is the stability of a repetition loop, which Section~\ref{sec:escape} treats. Why training stays at the threshold instead of crossing it is explained with a reduced model in Section~\ref{sec:spring}, and Experiment~2 measures the whole sequence on a transformer.

\subsection{Basins, fractal boundaries and slow transients}
\label{sec:transitions}

A stable state is only useful if the system reaches it. Which state it reaches depends on where it starts, and on the way there it can linger near states that are not stable at all.

\begin{figure}[t]
\centering
\includegraphics[width=\textwidth]{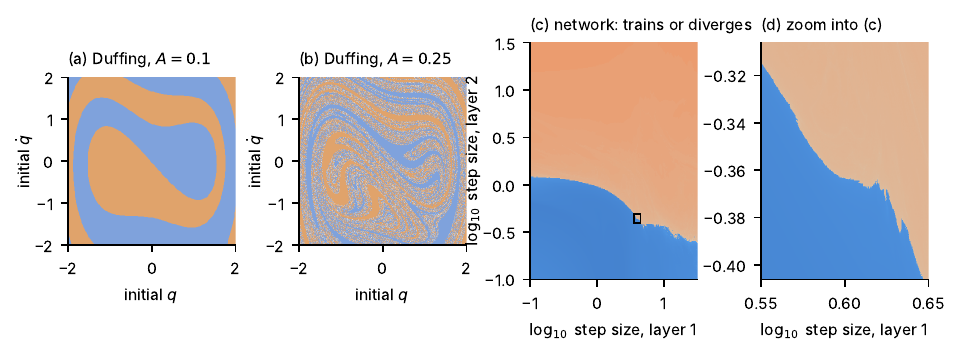}
\caption{Basins of attraction in a mechanical system and in training. (a, b) Forced double-well Duffing oscillator $\ddot q+0.25\dot q-q+q^3=A\cos(1.2\tau)$: the attractor reached from each initial condition after 40 forcing periods (blue: right well; orange: left well; gray: motion that keeps visiting both wells). At $A=0.1$ the basins are separated by a smooth boundary; at $A=0.25$ the boundary is fractal and a region of cross-well motion appears. (c) A one-hidden-layer tanh network (width 16) trained by full-batch gradient descent for 500 steps on eight random points, with separate step sizes for the two layers, following \citet{sohldickstein2024boundary}: blue runs train (darker: lower mean loss), orange runs diverge (lighter: later divergence). (d) Zoom into the box in (c).}
\label{fig:basins}
\end{figure}

\paragraph{Basins and metastability.} The basin of attraction of an attractor is the set of initial conditions that reach it. Basin boundaries can be fractal \citep{mcdonald1985fractal}. In a periodically forced oscillator, such as a ship rolling in beam seas, the basin of safe motion around the upright state erodes into fractal fingers as the wave amplitude grows, so that capsizing becomes sensitive to the initial state well before the safe attractor itself disappears \citep{thompson1990fractal,thompson1997designing}. The set of trainable hyperparameters of a neural network has the same kind of boundary: \citet{sohldickstein2024boundary} found that the boundary between step sizes that train and those that diverge is fractal, in the same way as the boundary of bounded orbits of the quadratic map that defines the Mandelbrot set. Figure~\ref{fig:basins} shows both cases side by side. For a forced double-well oscillator the two basins are separated by a smooth curve at small forcing and by interleaved, fractal bands at larger forcing. For a small network trained by gradient descent with one step size per layer, the boundary between training and divergence keeps its jagged structure when magnified 25 times. A metastable state holds trajectories for a very long time without being an attractor, for example a cluster configuration of tokens that eventually merges \citep{geshkovski2024dynamic}.

\paragraph{Slow transients and abrupt transitions.} Near a saddle, or near the ghost of a fixed point that has just disappeared in a saddle-node bifurcation, a trajectory slows down for a long time and then moves on quickly. Training curves show both patterns. High-dimensional loss surfaces have far more saddles than local minima \citep{dauphin2014identifying}, and for deep linear networks the learning dynamics can be solved exactly: modes of the input-output map are learned one after another, separated by plateaus whose length depends on the initialization \citep{saxe2013exact}. In recurrent networks, abrupt learning after a long plateau has been traced to the ghost of a saddle-node bifurcation \citep{dinc2025ghost}.

Abrupt transitions in what a model can do are the clock-2 counterpart of a bifurcation. Grokking, where test accuracy jumps long after the training loss has saturated, was first reported on algorithmic data \citep{power2022grokking}; inside the network the change is gradual, with circuit-level measures that improve steadily before the jump \citep{nanda2023progress}, and a solvable two-layer model shows a first-order phase transition \citep{rubin2023grokking}. In language models, induction heads form during a short window of training, with a visible bump in the loss curve \citep{olsson2022incontext}, and across model scale abilities can appear suddenly \citep{wei2022emergent}, although part of the effect comes from discontinuous metrics \citep{schaeffer2023emergent}. Further explanations of these transitions are reviewed in Appendix~\ref{app:related}.

\subsection{Synchronization: tokens that cluster over depth}
\label{sec:particles}

Basins and transients concern a single trajectory. Clock~1 is organized instead by many tokens acting on each other, and the matching concept from dynamics is synchronization.

\paragraph{Synchronization and consensus.} Weakly coupled oscillators can lock their phases. \citet{bennett2002huygens} reconstructed Huygens' 1665 observation of this effect: two pendulum clocks hanging from a common beam end up swinging in antiphase, coupled through the small motion of the beam. Metronomes on a freely rolling board do the same \citep{pantaleone2002synchronization}, and tidal friction has locked the Moon's rotation to its orbit, so it always shows the Earth the same face \citep{goldreich1966spin}. The Kuramoto model reduces many such systems to phases $\psi_i$ with natural frequencies $\omega_i$ and coupling strength $J$,
\begin{equation}
\dot\psi_i=\omega_i+\frac{J}{n}\sum_{j=1}^{n}\sin(\psi_j-\psi_i),
\label{eq:kuramoto}
\end{equation}
and shows a transition to collective synchrony once $J$ exceeds a critical value set by the spread of the $\omega_i$ \citep{kuramoto1975self,acebron2005kuramoto}. Attention over depth has the same structure without the natural frequencies. Each token drifts toward a softmax-weighted average of the others, with weights $e^{\beta\langle x_i,x_j\rangle}$, so aligned tokens pull harder on each other \citep{geshkovski2023mathematical}. On the circle, with $x_i=(\cos\psi_i,\sin\psi_i)$, this gives $\dot\psi_i=Z_i^{-1}\sum_j e^{\beta\cos(\psi_j-\psi_i)}\sin(\psi_j-\psi_i)$, a Kuramoto model whose coupling depends on alignment \citep{li2026diverse}. With $\beta=0$ every token ends in one cluster; with large $\beta$ several clusters form and survive over a long stretch of depth before merging (Section~\ref{sec:kuramoto}).

\begin{figure}[t]
\centering
\includegraphics[width=\textwidth]{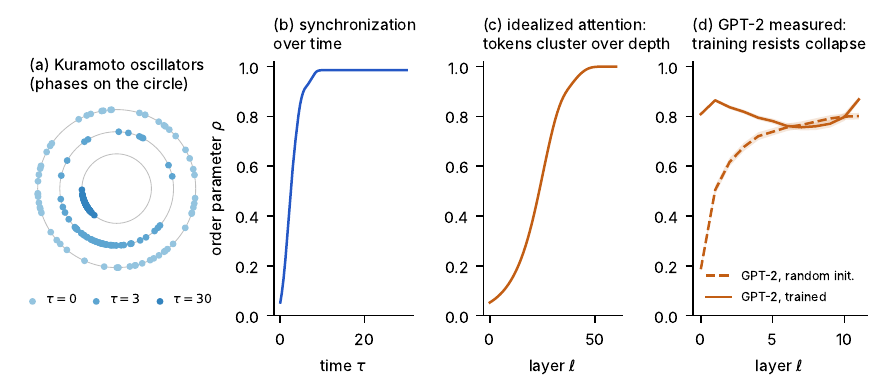}
\caption{Depth as a dynamical system: synchronization of oscillators and clustering of tokens. (a) Kuramoto model \eqref{eq:kuramoto} with $n=60$, $J=1.5$ and $\omega_i\sim\mathcal N(0,0.3^2)$: phases at $\tau=0$, 3 and 30 (outer to inner ring). (b) Its order parameter $\rho=|n^{-1}\sum_j e^{\mathrm{i}\psi_j}|$. (c) The attention dynamics on the circle with $\beta=1$ and 60 tokens: the same order parameter, now over layers. (d) GPT-2, measured: order parameter of the residual stream entering each block, $\rho_\ell=\|n^{-1}\sum_i x_i^\ell/\|x_i^\ell\|\|$, for 20 prompts (mean $\pm$ one standard deviation). With random weights the tokens synchronize over depth as the idealized model predicts ($\rho$ from 0.19 to 0.80). The trained model starts from aligned embeddings ($\rho=0.81$) and stays between 0.76 and 0.87 over the twelve blocks.}
\label{fig:pair_depth}
\end{figure}

\paragraph{On clock 1.} The transformer differs from a plain residual network because the update of token $i$ depends on all tokens. \citet{lu2019improving} read a transformer layer as a splitting scheme for a convection-diffusion equation of interacting particles. \citet{sander2021sinkformers} proved that attention with doubly stochastic (Sinkhorn) normalization is a Wasserstein gradient flow. \citet{geshkovski2023emergence,geshkovski2023mathematical} studied a model in which tokens live on the unit sphere (a stand-in for layer normalization) and follow
\begin{equation}
\dot x_i = \Pi_{x_i}\Big(\frac{1}{Z_i}\sum_{j} e^{\beta\langle W_Q x_i, W_K x_j\rangle} W_V x_j\Big),
\label{eq:particles}
\end{equation}
where $\Pi_x$ projects onto the tangent space, $W_Q,W_K,W_V$ are the query, key and value matrices, and $\beta$ acts as an inverse temperature. They proved that, for broad choices of $W_Q,W_K,W_V$, the tokens cluster as time (depth) grows, in the simplest cases to a single point. Clustering is the particle-system version of an older observation: stacks of pure attention lose rank doubly exponentially with depth \citep{dong2021attention}, and rank collapse at initialization slows training \citep{noci2022signal}. Similar results hold for oversmoothing in attention-based graph networks, which \citet{wu2023demystifying} analyze as nonlinear time-varying dynamical systems.

Real models have finite depth, so the long-time limit of \eqref{eq:particles} describes them only partly. \citet{geshkovski2024dynamic} proved that particles stay near a configuration of several clusters for an exponentially long time before they merge, and a stack of finite depth spends its layers inside such a long-lived transient.

\subsection{Noise-driven escape}
\label{sec:escape}
\label{sec:sgd}

Stable states and basins describe a system without noise. With noise, a system can leave a basin that it would never leave otherwise, and the time it takes is the subject of one of the oldest results of statistical physics.

\begin{figure}[t]
\centering
\includegraphics[width=\textwidth]{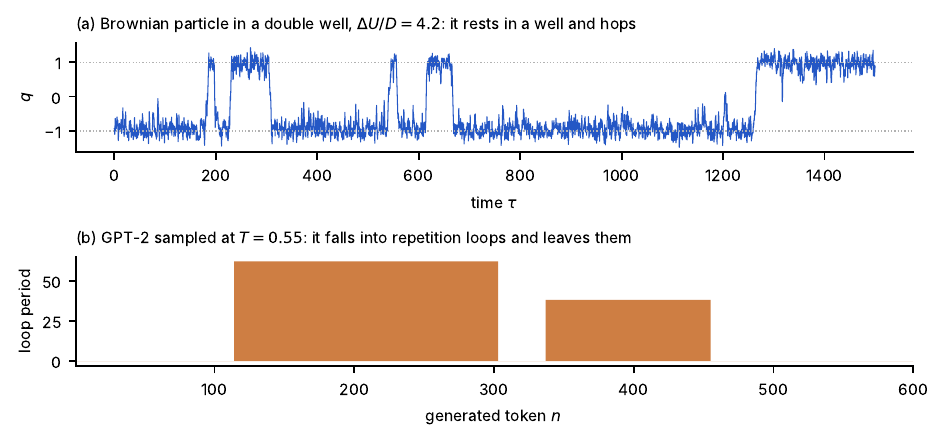}
\caption{Escape over time. (a) Position of an overdamped particle in the double well $U(q)=(q^2-1)^2/4$ under noise of intensity $D=0.06$: long rests near $q=\pm1$ separated by quick hops. (b) GPT-2 continuing one prompt with plain temperature sampling at $T=0.55$: at each generated token, the period of the repetition loop the text is in, where a stretch counts as a loop when it repeats at least three times (zero outside a loop). The text enters two loops, stays in each for more than a hundred tokens and leaves both.}
\label{fig:escape_time}
\end{figure}

\paragraph{Kramers' escape problem.} Add noise of intensity $D$ to a system resting at the bottom of a well of depth $\Delta U$. The mean time to leave the well grows as
\begin{equation}
\bar\tau\simeq\frac{2\pi}{\omega_{\mathrm{w}}\,\omega_{\mathrm{b}}}\,\zeta\,e^{\Delta U/D}
\label{eq:kramers}
\end{equation}
in the overdamped limit with friction coefficient $\zeta$, where $\omega_{\mathrm{w}}$ and $\omega_{\mathrm{b}}$ are set by the curvature at the bottom of the well and at the top of the barrier \citep{kramers1940brownian,hanggi1990reaction}. The exponential factor is the Arrhenius law for the rate of chemical reactions, with temperature in the role of $D$ \citep{arrhenius1889reaktionsgeschwindigkeit}. In engineering the same question decides whether a ship capsizes in random waves or a structure snaps through to a buckled state. Escape from potential wells under deterministic periodic forcing is a related problem, where the escape threshold in forcing amplitude depends on the forcing frequency and on the shape of the well \citep{farid2019escape,farid2022escape}. In mechanics escape is usually a failure to be avoided; in text generation the well is a repetition loop, and escape is what one wants. Figure~\ref{fig:escape_time} shows the two behaviors over time: a particle that rests in a well and hops, and GPT-2 sampled at a low temperature, which falls into repetition loops and leaves them.

\begin{figure}[t]
\centering
\includegraphics[width=0.8\textwidth]{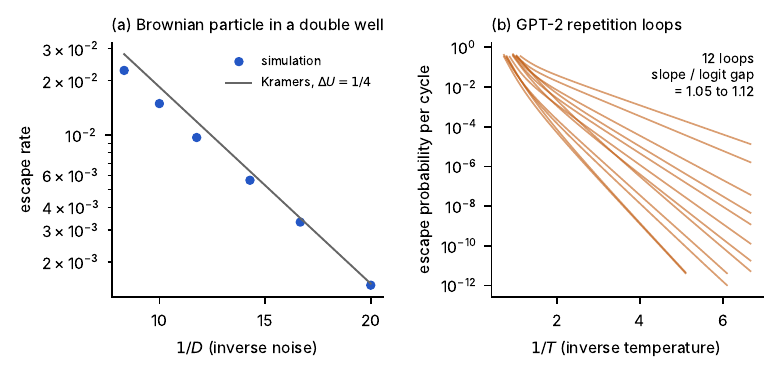}
\caption{Noise-driven escape in a mechanical system and in a language model, on the same axes. (a) An overdamped particle, $dq=-U'(q)\,d\tau+\sqrt{2D}\,d\mathcal{W}$, in the double well $U(q)=(q^2-1)^2/4$ with barrier $\Delta U=1/4$: inverse mean first-passage time from $q=-1$ to $q=+1$ (400 particles per noise level) against $1/D$, with Kramers' rate $(\sqrt2/2\pi)\,e^{-\Delta U/D}$. The fitted slope gives $\Delta U=0.23$. (b) Twelve repetition loops of GPT-2 from Experiment~3: exact probability of leaving the loop during one cycle, computed from the stored logits, against $1/T$. At low temperature each curve is a straight line whose slope is the logit gap of the loop, as Proposition~\ref{prop:arrhenius} predicts (ratio of slope to gap 1.05 to 1.12).}
\label{fig:pair_escape}
\end{figure}

\paragraph{On clock 2.} Minibatch noise turns clock~2 into a stochastic process. Stochastic modified equations approximate SGD with step size $\eta$ and batch size $B$ by the SDE
\begin{equation}
d\theta=-\nabla\mathcal{L}(\theta)\,dt+\sqrt{\tfrac{\eta}{B}\,C(\theta)}\;d\mathcal{W}_t ,
\label{eq:sme}
\end{equation}
where $C$ is the covariance of per-example gradients, $\mathcal{W}_t$ is a Wiener process, and time is measured in units of $\eta$ \citep{li2015stochastic}. Near a minimum with Hessian $H$ the drift is linear, the process is of Ornstein-Uhlenbeck type, and its stationary covariance $\Sigma$ solves the Lyapunov equation $H\Sigma+\Sigma H=(\eta/B)\,C$, so the ratio $\eta/B$ acts as a temperature \citep{mandt2017stochastic,smith2017bayesian,jastrzebski2017three}. Because SGD noise is anisotropic and state dependent, the stationary distribution is not a Gibbs distribution of the training loss, and the dynamics can converge to limit cycles instead of points \citep{chaudhari2017stochastic}.

Escape from a minimum is the most direct use of this theory. \citet{xie2020diffusion} derived a Kramers-type rate for SGD and found that, unlike isotropic Langevin dynamics, SGD favors flat minima exponentially in the ratio of barrier height to Hessian eigenvalues, and anisotropic noise aligned with the curvature helps escape from sharp minima \citep{zhu2018anisotropic}. Because the noise of SGD is multiplicative, the escape rate can instead scale as a power of the loss ratio between saddle and minimum \citep{mori2021powerlaw}, and heavy-tailed gradient noise points toward L\'evy-driven escape, dominated by single large jumps \citep{simsekli2019tailindex}. Which of these laws holds in language-model pretraining has not been measured.

\paragraph{On clock 3.} With greedy or low-temperature decoding, models fall into loops that repeat a phrase or sentence until the length limit \citep{holtzman2019curious}. At $T=0$ the chain is deterministic and the loop is one of its periodic orbits. \citet{xu2022learning} showed that loops reinforce themselves: the probability of repeating a sentence grows with the number of times it already appears in the context. Unlikelihood training \citep{welleck2019neural}, truncation rules \citep{hewitt2022truncation,meister2022locally} and feedback control of perplexity \citep{basu2020mirostat} can be read in part as attempts to keep the chain out of these attractors; \citet{fu2020theoretical} give a theoretical analysis of why they form. Recent work studies the inside of the loop: how it shows up in residual-stream dynamics before it becomes visible in the output \citep{zhang2026uncovering}, and the conditions under which a loop ``locks'' so that escape becomes impossible in practice under a given sampler \citep{chen2026repetition}. Loops also appear at the level of whole texts. Repeated paraphrasing converges to period-two attractor cycles even with sampling noise \citep{wang2025unveiling}. Reasoning models loop too, mostly at low temperature, which \citet{pipis2025wait} trace to errors in learning hard steps of a task and to temporally correlated errors.

Section~\ref{sec:loops} shows that for repetition loops the correspondence with Kramers' problem is an exact identity, and derives when a loop locks.

\subsection{Telling regimes apart from data}
\label{sec:regimes}

The previous subsections described the dynamics through models. In an experiment one usually has only a measured signal, and the same distinctions, between periodic motion, noise and chaos, must be read from it.

\begin{figure}[t]
\centering
\includegraphics[width=\textwidth]{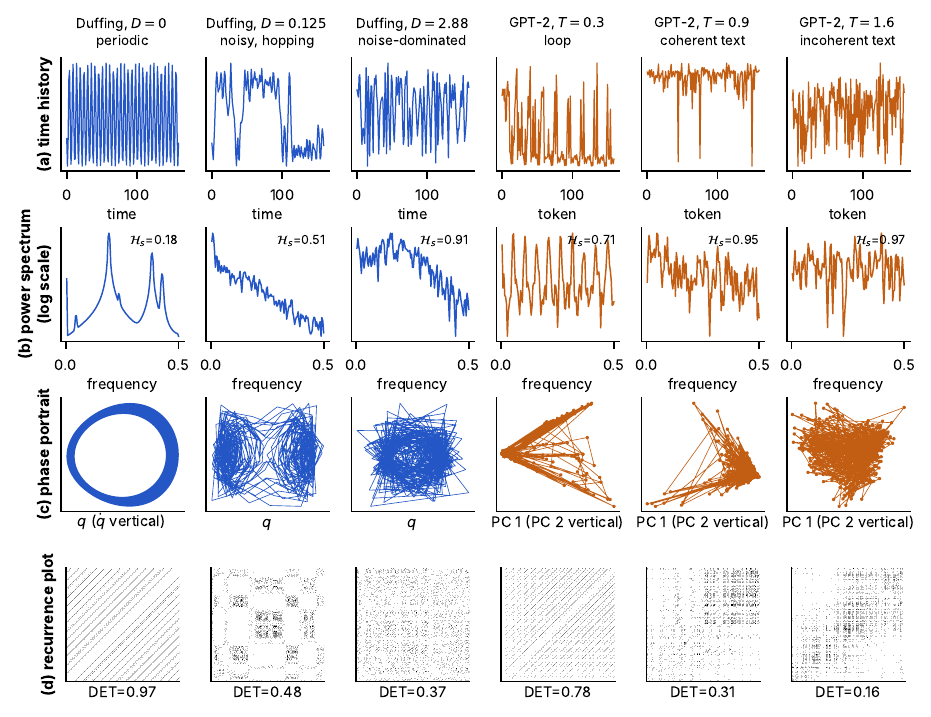}
\caption{Generation as a dynamical system: the same four pictures for a mechanical oscillator and for a language model. Left: the forced double-well Duffing oscillator $\ddot q+0.25\dot q-q+q^3=0.1\cos(1.2\tau)+\sqrt{0.5D}\,\xi(\tau)$, sampled once per time unit, at noise intensities $D=0$ (periodic orbit in one well), $0.125$ (noisy orbits with hops between the wells) and $2.88$ (noise-dominated). Right: GPT-2 continuing one prompt for 576 tokens at $T=0.3$ (a loop of period 19), $0.9$ (coherent text) and $1.6$ (incoherent text); the state is the final hidden state of each token, scaled to unit norm and standardized per feature, after a transient of 64 tokens. (a) Time history of the first coordinate ($q$, or the first principal coordinate of the hidden state). (b) Its power spectrum (Welch), with the spectral entropy $\mathcal{H}_s$. (c) Phase portrait: $(q,\dot q)$, or the first two principal coordinates. (d) Recurrence plot of the full state at a fixed recurrence rate of 5\%, with the determinism DET.}
\label{fig:atlas}
\end{figure}

\paragraph{Four pictures of a signal.} An experimentalist who sees only a signal classifies the motion with four standard pictures. The time history shows the motion directly. In the power spectrum a periodic motion gives sharp peaks at its frequency and harmonics, and noise or chaos gives a broad band. In the phase portrait a periodic orbit is a closed curve and noise fills a region. The recurrence plot marks the pairs of times $(i,j)$ at which the state returns close to itself, $R_{ij}=1$ if $\|u_i-u_j\|<\epsilon_R$, and a periodic orbit draws long lines parallel to the diagonal \citep{eckmann1987recurrence}. Two numbers condense the last two pictures: the normalized entropy of the spectrum, $\mathcal{H}_s$, from 0 for a single line to 1 for white noise, and the determinism DET, the fraction of recurrent points that lie on diagonal lines \citep{marwan2007recurrence}. Figure~\ref{fig:atlas} applies the four pictures to a forced double-well Duffing oscillator at three noise levels and to GPT-2 at three sampling temperatures. In both systems the quiet regime is a periodic orbit (for GPT-2, a repetition loop), with sharp spectral peaks and diagonal recurrence lines, and DET falls as the noise or the temperature grows. Over 10 noise seeds and 10 prompts, the median DET is 0.97, 0.48 and 0.42 for the Duffing oscillator and 0.67, 0.32 and 0.17 for GPT-2, and the median $\mathcal{H}_s$ is 0.18, 0.60 and 0.90 against 0.70, 0.95 and 0.95. The analogy has a limit that the pictures make plain: the spectra of coherent and of incoherent text are both close to flat, so for GPT-2 these two regimes differ in their recurrence structure and in the level of surprisal, not in the spectrum. Recurrence analysis of hidden states has recently been applied to the reasoning traces of DeepSeek-R1-Distill \citep{pham2026mind}, and recurrence plots and correlation dimensions to chain-of-thought trajectories \citep{jaca2026chaos}.

\paragraph{On clock 3.} At the level of statistics over many generations, temperature produces something close to a phase transition. \citet{nakaishi2024phase} varied $T$ in GPT-2 and found a low-temperature phase with repetitive structure, a high-temperature phase of incoherent text, and power-law correlations close to those of natural language at the transition point. Seen from clock~3, both phases are failures, and useful text lives in a window between them. The lower edge of the window is the loop escape of Section~\ref{sec:escape}. The upper edge, where sampling noise spreads over so much of the vocabulary that the text loses coherence, has no dynamical definition yet, and Experiment~3 locates it only through proxies. \citet{arnold2024phase} proposed statistical methods to detect such transitions automatically from output distributions. \citet{jones2026chaos} found that in the continuous feedback version of generation, temperature and activation steering move the system between periodic orbits and chaos with positive finite-time Lyapunov exponents, with period-doubling scaling and periodic orbits proved by interval arithmetic. In chain-of-thought trajectories, \citet{jaca2026chaos} report sensitive dependence on initial conditions together with bounded evolution and fractal structure in hidden-state space. Sorting many generated trajectories into regimes (fixed point, periodic, irregular) without labels is a response-regime identification problem, for which unsupervised data-driven methods exist in mechanics \citep{farid2024unsupervised}.

\subsection{Lyapunov exponents, order and chaos}
\label{sec:edgechaos}

The last concept measures how fast nearby trajectories separate. It is the quantity behind sensitivity to initial conditions, and it applies to all three clocks.

\paragraph{Lyapunov exponents.} For a trajectory $u_0,u_1,\dots$ the finite-time Lyapunov exponent over $K$ steps is
\begin{equation}
\Lambda_K(u_0) = \frac{1}{K}\log \sigma_{\max}\big(\mathrm{D}F(u_{K-1})\cdots \mathrm{D}F(u_0)\big),
\label{eq:ftle}
\end{equation}
where $\sigma_{\max}$ is the largest singular value. Positive values mean that nearby trajectories separate exponentially, so that a perturbation of size $\delta$ reaches size one after about $\log(1/\delta)/\Lambda_K$ steps. \citet{lorenz1963deterministic} found this sensitivity in a three-variable truncation of atmospheric convection, and it gives weather forecasts a finite horizon however accurate the initial data are. The limit $K\to\infty$, when it exists, is the largest Lyapunov exponent. Deep networks have a fixed and small number of steps, so only finite-time exponents make sense for clock~1 \citep{storm2023finitetime}.

\paragraph{On clock 1.} One line of work treats a randomly initialized deep network as a random dynamical system and asks whether small input differences grow or shrink with depth. \citet{poole2016exponential} found an order-to-chaos transition controlled by the weight and bias variances $\sigma_w^2,\sigma_b^2$. For a wide fully connected network with nonlinearity $\phi$, the correlation between two nearby inputs is multiplied at each layer by
\begin{equation}
\chi_1=\sigma_w^2\,\mathbb{E}\big[\phi'(\sqrt{Q^\ast}\,G)^2\big] ,
\label{eq:chi1}
\end{equation}
where $Q^\ast$ is the fixed point of the length map and $G$ is a standard Gaussian variable. Inputs converge for $\chi_1<1$ (order) and separate for $\chi_1>1$ (chaos), so $\ln\chi_1$ plays the role of a Lyapunov exponent per layer; \citet{schoenholz2016deep} showed that the depth over which signals propagate diverges at the transition, so networks are trainable to large depth only near it. The idea goes back to the chaotic phase of random recurrent networks \citep{sompolinsky1988chaos}. Requiring all singular values of the input-output Jacobian to be close to one (dynamical isometry) speeds up training further \citep{pennington2017resurrecting} and allowed 10{,}000-layer convolutional networks to train \citep{xiao2018dynamical}. For transformers, \citet{cowsik2024geometric} track the geometry of $n$ token representations as a discrete-time system of interacting particles and derive conditions on the initialization that make deep transformers trainable.

For trained networks, \citet{storm2023finitetime} computed finite-time Lyapunov exponents \eqref{eq:ftle} of fully connected classifiers and found ridges of large exponents in input space that separate the classes, much like coherent structures in fluid flows. Our first experiment (Section~\ref{sec:exp1}) applies the same quantity to the residual stream of language models across training.

\paragraph{What a dynamical model of depth must reproduce.} Interpretability research gives facts that any dynamical model of clock~1 has to reproduce. Predictions decoded from intermediate layers improve gradually, which suggests a trajectory that refines its estimate step by step \citep{belrose2023eliciting}. Deleting or swapping middle layers hurts much less than touching the first or last ones, consistent with distinct stages of the trajectory \citep{lad2024remarkable}. Token trajectories through depth stay close to a low-dimensional manifold \citep{sarfati2024lines}. A few activations are orders of magnitude larger than the rest and nearly independent of the input \citep{sun2024massive}, which together with attention sinks \citep{xiao2023efficient} can be read as a near fixed point of the depth dynamics. Experiment~1 measures how training changes the stretching along this trajectory.

\paragraph{On clock 3.} Training uses teacher forcing, while generation feeds the model its own outputs. Small errors therefore compound along clock~3, a problem known as exposure bias \citep{bengio2015scheduled,wang2020exposure}. Errors snowball in multi-step reasoning \citep{dziri2023faith} and in hallucinated justifications \citep{zhang2023language}, and next-token prediction can fail on planning tasks that require looking ahead \citep{bachmann2024pitfalls}. In dynamical terms these are statements about the growth of perturbations along a trajectory of the closed-loop system: if the finite-time exponent of clock~3 is positive, an early error of size $\delta$ grows roughly as $\delta\,e^{\Lambda_K K}$ over the next $K$ tokens. This exponent has not been measured for language models (\op{3}).

\paragraph{Koopman linearization.} Lyapunov exponents describe a nonlinear map through its linearization along one trajectory. A different, global linearization exists: a nonlinear map acts linearly on functions of the state, called observables, $h\mapsto h\circ F$ \citep{koopman1931hamiltonian}. Eigenfunctions of this Koopman operator give coordinates in which the dynamics is linear, and finite approximations can be learned from data \citep{mezic2005spectral}. Dynamic mode decomposition, for instance, extracts the oscillation frequencies and spatial modes of a jet in crossflow directly from snapshots \citep{rowley2009spectral}. Related data-driven methods learn linear embeddings of nonlinear dynamics with neural networks \citep{lusch2017deep} or sparse governing equations from data \citep{brunton2015discovering}. Koopman methods have also been applied to the training dynamics of neural networks \citep{dogra2020optimizing,redman2021operator}.

\section{Why the analogies hold: reduced-order models}
\label{sec:reduced}

Engineers often understand a complex system through a simple model that shares its governing terms. Sloshing in a partially filled liquid tank is a free-surface flow, yet its response regimes, including the strongly nonlinear ones with hydraulic impacts on the walls, are reproduced by an equivalent mechanical model in which a linear oscillator is coupled to a vibro-impact unit \citep{farid2017response}. Section~\ref{sec:dictionary} showed that the clocks and textbook physical systems behave alike. This section explains why, with reductions of the same kind. For each one we name the physical model and its governing equation (Figure~\ref{fig:models}), derive the part of the language model that reduces to it, and state where the reduction stops holding. Table~\ref{tab:reduced} summarizes the six cases, and the subsections follow its order: three results on clock~3, two on clock~2 and one on clock~1.

\begin{table}[t]
\centering
\small
\caption{Reduced-order models of language-model behavior. The shared term is the part of the governing equation that produces the behavior in both systems.}
\label{tab:reduced}
\begin{tabular}{p{0.2\linewidth}p{0.22\linewidth}p{0.26\linewidth}p{0.21\linewidth}}
\toprule
Language model & Reduced model & Shared term & Status \\
\midrule
Escape from a loop (clock 3) & Particle in a well under noise, Fig.~\ref{fig:models}(b) & Boltzmann factor: $e^{-\Delta/T}$ and $e^{-\Delta U/D}$ & Exact identity (Prop.~\ref{prop:arrhenius}) \\
Loops that hold for a while (clock 3) & Particle in a deepening well, Fig.~\ref{fig:models}(d) & Barrier that grows with residence time & Fitted; $T_{1/2}$ within 9\% to 25\% \\
Regimes of generation (clock 3) & Noisy double-well Duffing oscillator, Fig.~\ref{fig:models}(a) & Ratio of barrier to noise & Same ordering, measured \\
Period two at the edge of stability (clock 2) & Explicit Euler step on a nonlinear spring & Step multiplier $1-\eta\lambda$ crossing $-1$, cubic term of the loss & Derived; multi-mode edge not covered \\
Loss spike (clock 2) & Ball in a valley whose stiffness changes along it, Fig.~\ref{fig:catapult} & Step multiplier $1-\eta\lambda$ below $-1$ where the valley is stiff & Exact for the two-parameter model; untested at scale \\
Clustering over depth (clock 1) & Kuramoto oscillators, Fig.~\ref{fig:models}(c) & Coupling $\sin(\psi_j-\psi_i)$ & Exact for idealized attention; trained GPT-2 departs \\
\bottomrule
\end{tabular}
\end{table}

\subsection{Escape from a loop: the softmax is a Boltzmann distribution}
\label{sec:loops}

Repetition loops are the case where the reduction is exact, so we start with the noise-driven escape of Section~\ref{sec:escape}.

\paragraph{The softmax is a Boltzmann distribution.} In statistical mechanics a system in contact with a heat bath occupies a state of energy $E$ with probability proportional to $e^{-E/k_BT}$. The sampler \eqref{eq:clock3} has exactly this form, with the energy of token $y$ equal to minus its logit, $E_y=-z_y$, and $k_B=1$. A token is a state, a logit is minus an energy, and the sampling temperature is a temperature in the physical sense. Suppose the chain has entered a loop of period $k$ with tokens $a_1,\dots,a_k$. As long as the sampler keeps choosing the loop token, the context is deterministic, so the logits $z_i$ at the $i$-th step of the loop are fixed numbers. The loop token $a_i$ has the highest logit, and every other token $y$ is an exit whose energy is higher by the logit gap $\Delta_i^{y}=z_i(a_i)-z_i(y)>0$. The probability of staying in the loop at step $i$ is
\begin{equation}
p_i(T)=\frac{1}{1+S_i(T)},\qquad S_i(T)=\sum_{y\neq a_i} e^{-\Delta_i^{y}/T},
\label{eq:stay}
\end{equation}
and the probability of leaving during one cycle is $r(T)=1-\prod_{i=1}^{k}p_i(T)$ (Figure~\ref{fig:models}b). At low temperature the probability of leaving at step $i$ is
\begin{equation}
1-p_i(T)=\frac{S_i(T)}{1+S_i(T)}\;\approx\;m_i\,e^{-\Delta_i/T}\qquad (T\to0),
\label{eq:boltz_exit}
\end{equation}
where $\Delta_i$ is the smallest gap and $m_i$ the number of tokens that attain it. This is the Arrhenius factor of Kramers' problem \eqref{eq:kramers}, with the logit gap in place of the barrier height $\Delta U$ and $T$ in place of the noise intensity $D$ (Figure~\ref{fig:models}b). The correspondence is an identity, which is why every loop in Figure~\ref{fig:pair_escape}(b) gives a straight line whose slope is its gap. Two differences remain. A Brownian particle diffuses up to the barrier, and the curvatures at the bottom and the top of the well set Kramers' prefactor; the sampler jumps in one step, so the prefactor is the number of exits $m_i$. In addition, the energies of a language model change as the context grows, which Section~\ref{sec:deepening} treats.

\begin{proposition}[Arrhenius law for loop escape]
\label{prop:arrhenius}
Let $\Delta_{\min}=\min_i\min_{y\neq a_i}\Delta_i^{y}$ and let $m$ be the number of pairs $(i,y)$ that attain it. Then $r(T)\le \sum_{i}S_i(T)$ for all $T>0$, and
\begin{equation}
r(T)= m\, e^{-\Delta_{\min}/T}\,\big(1+o(1)\big)\quad\text{as } T\to0,\qquad\text{so}\qquad \lim_{T\to0}T\log r(T)=-\Delta_{\min}.
\end{equation}
\end{proposition}
\noindent The proof is in Appendix~\ref{app:proofs}.

The logit gap therefore plays the role of the barrier height $\Delta U$ in Kramers' formula \citep{kramers1940brownian}, and $T$ plays the role of noise intensity. Proposition~\ref{prop:arrhenius} follows almost directly from the Gibbs form of the softmax. \citet{chen2026repetition} use the same product of per-step continuation probabilities to ask when a sampler locks a loop exactly, with probability one, and show that plain temperature sampling never does. Here a loop locks in a weaker, probabilistic sense: the probability of never escaping is positive (Proposition~\ref{prop:lock}), or escape takes longer than the generation (Section~\ref{sec:exp3}).

\subsection{Loops that hold for a while: a well that deepens while it is occupied}
\label{sec:deepening}

\begin{figure}[t]
\centering
\includegraphics[width=\textwidth]{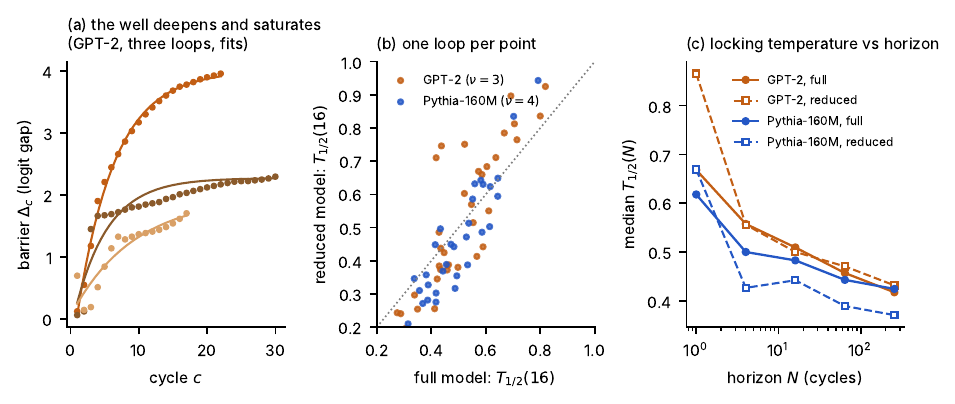}
\caption{The reduced-order model of loop escape against the full computation. (a) Smallest logit gap of each cycle (dots) along three GPT-2 loops of at least 15 cycles, with the saturating fit of \eqref{eq:reduced} (lines). (b) Locking temperature for a horizon of 16 cycles after a burn-in of two cycles, from the reduced model (three numbers per loop read from the greedy run, one attempt number $\nu$ per model) against the full computation from all logits, one point per loop. (c) Median locking temperature against the horizon for both models, full (filled) and reduced (open).}
\label{fig:reduced}
\end{figure}

The Boltzmann identity \eqref{eq:boltz_exit} holds at every step, but the energies change from cycle to cycle, because each completed cycle adds the same text to the context and the model grows more confident that it will repeat it, the self-reinforcement reported by \citet{xu2022learning}. Figure~\ref{fig:reduced}(a) shows the smallest gap $\Delta_c$ of each cycle along three GPT-2 loops: it grows and then saturates. The physical counterpart is a particle in a well that deepens while it is occupied (Figure~\ref{fig:models}d). Rock friction behaves this way: the static friction between two surfaces grows with the time they have been in contact \citep{dieterich1972time}, so a block that has rested longer needs a larger force to slide.

To see what this growth does to escape, let $\Delta_c$ be the smallest gap during cycle $c$ and $r_c(T)$ the escape probability during that cycle. The probability of never leaving is $\prod_{c\ge1}(1-r_c(T))$, which is positive exactly when $\sum_c r_c(T)<\infty$ (given $r_c<1$).

\begin{proposition}[Deepening wells lock]
\label{prop:lock}
If $\Delta_c\ge\Delta_1+b\,(c-1)$ for some $b>0$, then for every $T>0$ the probability of never escaping is at least $1-k(|\mathcal V|-1)\,e^{-\Delta_1/T}/(1-e^{-b/T})$ whenever this bound is positive, and the probability itself is positive for every $T>0$. If instead the logits along the loop converge from cycle to cycle, so that $\Delta_c\to\Delta_\infty<\infty$, escape eventually happens with probability one, but the expected number of cycles before escape grows like $e^{\Delta_\infty/T}$ as $T\to0$.
\end{proposition}
\noindent The proof is in Appendix~\ref{app:proofs}.

Between linear growth and saturation lies a growth law with a sharp threshold in temperature.

\begin{proposition}[Locking temperature]
\label{prop:tlock}
If $\Delta_c=\Delta_1+\gamma\ln c$ for all $c\ge1$ with $\gamma>0$, the probability of never escaping is positive for $T<\gamma$, and escape happens with probability one for $T\ge\gamma$. For $T>\gamma$ the probability of staying in the loop for $N$ cycles is $\exp\!\big(-\Theta(N^{1-\gamma/T})\big)$.
\end{proposition}
\noindent The proof is in Appendix~\ref{app:proofs}.

Proposition~\ref{prop:tlock} gives an infinite-horizon locking temperature $T_{\mathrm{lock}}=\gamma$, below which survival decays to a positive constant and above which it decays as a stretched exponential. Linear growth locks at every temperature and saturating growth at none, so the three cases of Propositions~\ref{prop:lock} and \ref{prop:tlock} cover the possible outcomes for an unbounded generation. A real generation has a finite budget of $N$ cycles. If the barrier has saturated at $\Delta_\infty$ and the escape probability per cycle is $m\,e^{-\Delta_\infty/T}\ll1$, survival over $N$ cycles equals one half at
\begin{equation}
T_{1/2}(N)\approx\frac{\Delta_\infty}{\ln\!\big(mN/\ln2\big)},
\label{eq:thalf}
\end{equation}
so the locking temperature observed in a saturating loop falls like $1/\ln N$ with the length of the generation. In practice logits are bounded, so the gaps cannot grow without limit. Experiment~3 (Section~\ref{sec:exp3}) measures $\Delta_c$ along real loops of GPT-2 and Pythia, tests Propositions~\ref{prop:arrhenius} and \ref{prop:lock} against the exact survival probabilities and against Monte Carlo sampling, and finds that the barriers saturate. A reduced model with a saturating barrier is therefore the right one for real loops.

\paragraph{Truncated sampling turns the probabilistic lock into an exact one.} Deployed models often sample with top-$p$ truncation \citep{holtzman2019curious}. Once the probability of the repeated token reaches $p$, the nucleus holds that token alone and the loop locks exactly, the case analyzed by \citet{chen2026repetition}. With a saturated gap $\Delta_\infty$ to the closest rival, the repeated token has probability at most $1/(1+e^{-\Delta_\infty/T})$, so for $p>1/2$ exact locking under top-$p$ requires
\begin{equation}
T\le\frac{\Delta_\infty}{\ln\!\big(p/(1-p)\big)}.
\label{eq:ttopp}
\end{equation}
Other rivals lower the probability of the repeated token further, so Eq.~\eqref{eq:ttopp} is an upper bound on the temperature below which top-$p$ locks a saturated loop exactly. Above it the lock stays probabilistic and Eq.~\eqref{eq:thalf} applies. We thank Yichuan Chen for this observation.

\paragraph{A reduced model for real loops.} The reduced model has two ingredients, a saturating barrier fitted to the gaps of each loop,
\begin{equation}
\Delta_c=\Delta_\infty-(\Delta_\infty-\Delta_1)\,e^{-(c-1)/c_\ast},\qquad r_c(T)=1-\big(1+e^{-\Delta_c/T}\big)^{-\nu},
\label{eq:reduced}
\end{equation}
and an escape probability per cycle $r_c$ for a single exit attempted $\nu$ times per cycle. The three numbers $\Delta_1,\Delta_\infty,c_\ast$ of each loop come from its greedy run alone, and $\nu$ plays the role of Kramers' prefactor; we use one value per model, chosen on a grid ($\nu=3$ for GPT-2, $\nu=4$ for Pythia-160M). Figure~\ref{fig:reduced} compares the locking temperature $T_{1/2}(N)$ of this model with the full computation from all logits of all steps (Section~\ref{sec:exp3}). For horizons of 4 to 256 cycles the reduced model is within a median of 9\% to 25\% of the full value for individual loops, with correlations of 0.74 to 0.95 across loops, and it reproduces the slow fall of $T_{1/2}$ with $N$. For a single cycle it is off by 30\% to 50\%. For long horizons the reduced model gives $T_{1/2}(N)\approx\Delta_\infty/\ln(\nu N/\ln2)$, the form of \eqref{eq:thalf}, since an escape within $N$ cycles becomes likely once $\nu N e^{-\Delta_\infty/T}$ is of order one, so the locking temperature falls like $1/\ln N$.

\subsection{One number orders the regimes of generation}
\label{sec:collapse}

\begin{figure}[t]
\centering
\includegraphics[width=0.85\textwidth]{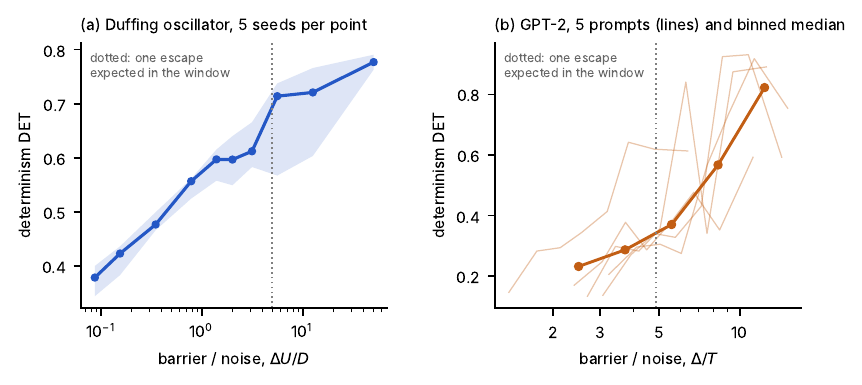}
\caption{Determinism of the trajectory against the ratio of barrier height to noise intensity. (a) Forced double-well Duffing oscillator of Figure~\ref{fig:atlas}, ten noise levels, five seeds each (median and range), with $\Delta U=1/4$. (b) GPT-2, five prompts whose greedy runs end in a loop, sampled at eight temperatures from 0.3 to 1.4; $\Delta$ is the median logit gap of the last three cycles of the greedy loop. Thin lines: individual prompts; thick line: median in bins of $\Delta/T$. Dotted lines: the ratio at which one escape is expected within the observed window, $\ln(N/\ln2)$ for the oscillator and $\ln(\nu N/\ln2)$ with $\nu=3$ for GPT-2.}
\label{fig:collapse}
\end{figure}

The Boltzmann identity also accounts for the regimes of Section~\ref{sec:regimes}. If escape from a well is governed by the factor $e^{-\text{barrier}/\text{noise}}$, the motion of a system with wells is organized by one dimensionless number, the ratio of barrier height to noise intensity, whatever the other details of the system. The prediction is that the oscillator and the language model of Figure~\ref{fig:atlas} leave their periodic orbits at the same value of this ratio, a value set by the length of the observation. For the double-well Duffing oscillator with unit mass, friction coefficient $\zeta$ and noise term $\sqrt{2\zeta D}\,\xi$, the unforced motion has the stationary density $\propto\exp\{-(\dot q^2/2+U(q))/D\}$, so the depth of a well in units of the noise is $\Delta U/D$, with $\Delta U=1/4$ for $U=-q^2/2+q^4/4$. If an escape has probability of order $e^{-\Delta U/D}$ per period, the oscillator is expected to leave a well within $N$ periods once $\Delta U/D\lesssim\ln(N/\ln2)$. Above this value it stays on a periodic orbit in one well, close to it the oscillator hops between wells, and far below it the noise dominates. For a loop of a language model the corresponding ratio is $\Delta/T$, with $\Delta$ the barrier of the loop, and the same estimate gives $\Delta/T\lesssim\ln(\nu N/\ln2)$, with $N$ the number of cycles in the window. Figure~\ref{fig:collapse} plots the determinism of the trajectory against this ratio for both systems. For the oscillator (512 samples, about 98 forcing periods, so $\ln(N/\ln2)\approx4.9$) DET rises from 0.38 to 0.78 as $\Delta U/D$ grows from 0.09 to 50, with its largest step between 3.1 and 5.6. For GPT-2 we sampled five prompts whose greedy runs loop, at eight temperatures from 0.3 to 1.4, and took $\Delta$ as the gap of the last cycles of the greedy loop. DET rises with $\Delta/T$ (correlation 0.75 with $\log(\Delta/T)$ over 40 runs), and the binned median climbs from 0.37 to 0.82 between $\Delta/T\approx5.6$ and 12, just above the value 4.9 at which one escape is expected in the window. The scatter between prompts is large, and five prompts do not determine the shape of the curve, but both systems leave the periodic regime when the barrier falls to a few times the noise, at the value set by the length of the observation.

\subsection{The edge of stability: an Euler step on a nonlinear spring}
\label{sec:spring}

We now turn to clock~2. The edge of stability of Section~\ref{sec:eos}, where training settles on a period-2 orbit at the threshold $2/\eta$, has a reduction that takes one line. Gradient descent $\theta_{t+1}=\theta_t-\eta\nabla\mathcal L(\theta_t)$ is the explicit Euler scheme with step $\eta$ for the gradient flow $\dot\theta=-\nabla\mathcal L(\theta)$, the motion of an overdamped particle in the potential $\mathcal L$. Restrict it to the eigenvector $v$ of the largest Hessian eigenvalue $\lambda$ at a minimum $\theta^\ast$, with $q=\langle\theta-\theta^\ast,v\rangle$ and $\mathcal L_v(q)=\mathcal L(\theta^\ast+qv)$. Then
\begin{equation}
q_{t+1}=q_t-\eta\,\mathcal L_v'(q_t),\qquad \mathcal L_v'(q)=\lambda q+\tfrac12\mathcal L_v'''q^2+\tfrac16\mathcal L_v''''q^3+\dots,
\label{eq:gd_spring}
\end{equation}
and $-\mathcal L_v'(q)$ is the restoring force of a spring with linear, quadratic and cubic stiffness, the Duffing spring of Figure~\ref{fig:models}(a) when the quadratic term vanishes. The linear term alone gives $q_{t+1}=(1-\eta\lambda)q_t$, whose multiplier crosses $-1$ at $\eta\lambda=2$. The minimum then loses stability through a period-doubling bifurcation, the same instability that limits the time step of the explicit Euler scheme for a stiff spring. The nonlinear terms decide what follows. For the double well $\mathcal L_v(q)=(q^2-1)^2/4$ around $q=1$, where $\lambda=2$, the iterate settles on a period-2 orbit for $\eta\lambda$ between 2 and about 2.47, on a period-4 orbit from 2.48, on a period-8 orbit near 2.58, and on no orbit of period up to 64 from about 2.62 (Figure~\ref{fig:pair_training}a,b). A period-2 orbit is what produces the spectral line at frequency $1/2$.

\paragraph{A spring whose stiffness responds to the motion.} A network adds a third ingredient: $\lambda$ depends on $\theta$, so the oscillation along $v$ changes the curvature that drives it, and the motion and the stiffness are coupled. \citet{damian2022selfstabilization} reduce this feedback to a two-variable map. With $q_t$ the displacement along the top eigenvector $v$ and $\lambda_t$ the sharpness, linearized around a point where it equals $2/\eta$, the two evolve as
\begin{equation}
q_{t+1}=(1-\eta\lambda_t)\,q_t,\qquad \lambda_{t+1}=\lambda_t+\eta\big(\alpha-\|\nabla\lambda_{\max}\|^2 q_t^2\big),
\label{eq:selfstab}
\end{equation}
where $\alpha>0$ measures progressive sharpening and the cubic term of the loss enters through the gradient of the sharpness, $\nabla\lambda_{\max}$. The first equation is the linear step of \eqref{eq:gd_spring} with a stiffness that now moves. The oscillation grows while $\eta\lambda_t>2$, and its square pulls $\lambda_t$ back, so the pair cycles around the threshold instead of diverging. Mechanically, this is a mass in an elastic well whose walls soften when the mass swings hard, an oscillator whose stiffness is a slow variable driven by the amplitude of the motion. The feedback explains why the transformer of Experiment~2 shows a line at $1/2$ (Figure~\ref{fig:pair_training}c,f) across two decades of step size instead of a cascade to chaos at one step size: training keeps its own bifurcation parameter at the threshold. The one-direction reduction holds while one eigenvalue dominates. In our transformer two to six eigenvalues sit above $2/\eta$ (Section~\ref{sec:exp2}), which neither \eqref{eq:gd_spring} nor the two-variable map describes. Related results include conditions under which gradient descent converges while being locally unstable \citep{arora2022gradient,ahn2022unstable}, solvable minimal models with edge-of-stability dynamics \citep{zhu2022edgeofstability}, convergence to stable period-two orbits beyond the threshold, a period-doubling bifurcation in the strict sense \citep{chen2022edge}, and the implicit penalty on the gradient norm that backward error analysis finds in the finite step \citep{barrett2020implicit}.

\subsection{A loss spike: a ball thrown out of a stiff valley}
\label{sec:catapult}

The same instability, seen in a loss landscape whose stiffness changes from place to place, produces loss spikes. Large pretraining runs show such spikes. PaLM handled them by restarting from an earlier checkpoint and skipping data \citep{chowdhery2022palm}. \citet{wortsman2023smallscale} reproduced two instability mechanisms at small scale, growth of attention logits and divergence of the output logits, and studied how they depend on learning rate. Attention entropy collapse is another proposed cause \citep{zhai2023stabilizing}, as is the behavior of the Adam update at large scale \citep{molybog2023theory}; conditions on the scale of embeddings and residual connections can prevent spikes \citep{takase2023spike}. The smallest model with a loss spike has two parameters, and it shows the mechanism in a form that can be drawn.

\begin{figure}[t]
\centering
\includegraphics[width=\textwidth]{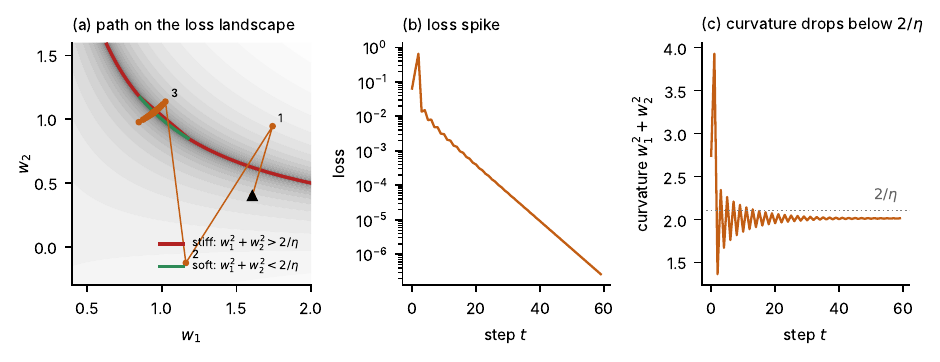}
\caption{A loss spike in two parameters, seen as a ball thrown out of a stiff valley. Gradient descent on $(w_1w_2-1)^2/2$ with $\eta=0.95$, started near $(w_1,w_2)=(1.6,0.4)$ where the curvature across the valley of minima, $w_1^2+w_2^2=2.74$, exceeds $2/\eta=2.11$. (a) Path of the iterate (triangle: start; numbers: the first three steps) on the loss landscape (gray: log loss); the valley of minima $w_1w_2=1$ is red where it is too stiff for the step size and green where it is not. (b) The loss jumps from 0.064 to 0.66 and then decays. (c) The curvature oscillates and settles at 2.01, below $2/\eta$.}
\label{fig:catapult}
\end{figure}

In the catapult model of \citet{lewkowycz2020large} the product $w_1w_2$ of two scalar weights is fit to a single target with step size $\eta$. The loss $(w_1w_2-1)^2/2$ has a valley of minima along the hyperbola $w_1w_2=1$, and the curvature across the valley at $(w_1,w_2)$ is $w_1^2+w_2^2$, large far out along the axes and smallest at $w_1=w_2=1$ (Figure~\ref{fig:catapult}a). The mechanical picture is a ball moved by the explicit Euler steps of Section~\ref{sec:spring} in a channel whose walls are steep at one end and gentle at the other. Started on the steep part, where $\eta(w_1^2+w_2^2)>2$, each step overshoots the floor and the ball bounces from wall to wall with growing amplitude, the period-doubling instability of \eqref{eq:gd_spring}: the loss spikes. The bouncing also carries the ball along the channel toward the gentle end, and once it reaches the part where $\eta(w_1^2+w_2^2)<2$, the steps stop overshooting and the ball settles on the floor. With $\eta=0.95$ the loss rises tenfold within two steps and the iterate settles where the curvature, 2.01, is below $2/\eta=2.11$ (Figure~\ref{fig:catapult}b,c). \citet{lewkowycz2020large} call this the catapult phase; above a second threshold training diverges, and \citet{kalra2023phase} mapped these early-training regimes as a function of learning rate, depth and width. Whether the spikes of large pretraining runs are catapults, crossings of the stability boundary or escapes from a basin is open (\op{4}).

\subsection{Clustering over depth: attention on the circle is a Kuramoto model}
\label{sec:kuramoto}

The last reduction concerns clock~1 and explains the synchronization of Section~\ref{sec:particles}. In the transformer model of \citet{geshkovski2023mathematical}, layer normalization keeps every token on the unit sphere, and self-attention with identity query, key and value matrices moves token $i$ by
\begin{equation*}
\dot x_i=\Pi_{x_i}\Big(Z_i^{-1}\sum_j e^{\beta\langle x_i,x_j\rangle}x_j\Big),\qquad Z_i=\sum_j e^{\beta\langle x_i,x_j\rangle},
\end{equation*}
where $\Pi_x$ projects onto the tangent space of the sphere at $x$. In two dimensions write $x_i=(\cos\psi_i,\sin\psi_i)$. The unit tangent at $x_i$ is $(-\sin\psi_i,\cos\psi_i)$, its inner product with $x_j$ is $\sin(\psi_j-\psi_i)$, and $\langle x_i,x_j\rangle=\cos(\psi_j-\psi_i)$, so
\begin{equation}
\dot\psi_i=Z_i^{-1}\sum_j e^{\beta\cos(\psi_j-\psi_i)}\sin(\psi_j-\psi_i).
\label{eq:attn_kuramoto}
\end{equation}
For $\beta=0$, $Z_i=n$ and \eqref{eq:attn_kuramoto} is the Kuramoto model \eqref{eq:kuramoto} with $\omega_i=0$ and $J=1$, the model of identical metronomes on a common board, which synchronize from almost every initial condition (Figure~\ref{fig:models}c). In both systems each unit is pulled toward a weighted average of the others, and for $\beta>0$ the pull is stronger between tokens that are already close, which lets several clusters form and persist (Section~\ref{sec:particles}). The reduction accounts for Figure~\ref{fig:pair_depth}(c) and for GPT-2 with random weights in panel (d). Trained GPT-2 departs from it: its order parameter stays near 0.8 over all twelve blocks. The reduction drops the learned query, key and value maps, the MLP blocks and the finite depth. One candidate explanation is that these terms act like the spread of natural frequencies $\omega_i$ in the Kuramoto model, which opposes synchronization; we have not tested it.

\section{Three experiments, one per clock}
\label{sec:experiments}

Sections~\ref{sec:dictionary} and~\ref{sec:reduced} described the three clocks with the concepts of nonlinear dynamics and explained several correspondences with reduced models. This section tests three of them with new measurements on open models, one per clock. Experiment~1 asks how training changes the finite-time Lyapunov exponents of the forward pass (Section~\ref{sec:edgechaos}). Experiment~2 maps how gradient descent on a transformer moves through the regimes of Section~\ref{sec:eos} as the step size grows. Experiment~3 tests the escape theory of Sections~\ref{sec:loops} and~\ref{sec:deepening} on the repetition loops of GPT-2 and Pythia, and follows the coherent window it implies across model size and into a reasoning model. Every run fits on a two-core CPU; most take under an hour, and the largest models take a few hours. Details are in Appendix~\ref{app:details}, and the code reproduces every number.

\subsection{Experiment 1: the Lyapunov profile of the residual stream over training (clocks 1 and 2)}
\label{sec:exp1}

\paragraph{Question and setup.} Finite-time Lyapunov exponents \eqref{eq:ftle} describe how a forward pass stretches perturbations, layer by layer. How does this profile change as the model trains? This question couples clock~1 to clock~2 (\op{1}). We use the public training checkpoints of Pythia \citep{biderman2023pythia}: eleven checkpoints of Pythia-70M (6 layers), from initialization to step 143{,}000, and six checkpoints of Pythia-160M (12 layers). Each block is treated as a map $F_\ell$ on the full residual stream of a 16-token context. For each block we estimate the largest singular value $\sigma_\ell$ of its Jacobian by power iteration and report the relative exponent
\begin{equation}
\tilde\Lambda_\ell=\log\sigma_\ell-\log\big(\|x^{\ell+1}\|/\|x^{\ell}\|\big),
\end{equation}
which measures stretching beyond the growth of the residual norm. For the whole stack we report $\Lambda=\frac{1}{L}\log\sigma_{\max}(\mathrm{D}F_{0\to L})$, and the ratio $\kappa=\log\sigma_{\max}(\mathrm{D}F_{0\to L})/\sum_\ell\log\sigma_\ell$. Since $\sigma_{\max}$ of a product is at most the product of the $\sigma_\ell$, $\kappa\le1$ whenever $\sum_\ell\log\sigma_\ell>0$, which holds throughout our measurements, and $\kappa$ near one means that the most expanded direction of each layer is passed on to the next. Values are means over eight prompts, and $\kappa$ is the ratio of the two means.

\begin{figure}[t]
\centering
\includegraphics[width=\textwidth]{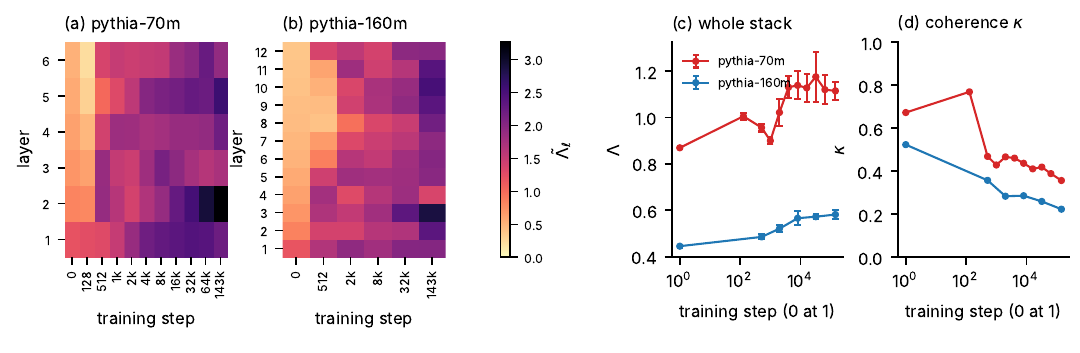}
\caption{Experiment 1: finite-time Lyapunov exponents of the residual stream across public Pythia checkpoints. (a, b) Relative per-layer exponent $\tilde\Lambda_\ell$ (stretching of the top singular direction beyond the growth of the residual norm) for each layer and checkpoint. (c) Depth-averaged exponent of the whole stack, $\Lambda$; bars: standard deviation over prompts. (d) Coherence $\kappa$ between per-layer and whole-stack stretching; values below one mean that the most expanded directions of consecutive layers are not aligned.}
\label{fig:exp1}
\end{figure}

\paragraph{Results.} Figure~\ref{fig:exp1} shows three things. First, at initialization both models have the same profile, decreasing with depth ($\tilde\Lambda_\ell$ from 1.20 to 0.53 in Pythia-70M and from 1.19 to 0.39 in Pythia-160M), which reflects the shared initialization scheme. Second, training makes every layer more expansive. The mean of $\tilde\Lambda_\ell$ over layers rises from 0.75 to 2.32 in Pythia-70M and from 0.59 to 2.17 in Pythia-160M, and it is still rising at the last checkpoint. The rise is not monotone at first: at step 128 the deepest layers of Pythia-70M contract further (layer 6 drops from 0.53 to 0.22) before all layers expand. Third, the whole stack does not follow. In Pythia-70M, $\Lambda$ goes from 0.87 to about 1.1 and stays between 1.11 and 1.18 from step 4{,}000 onward, which is under 3\% of training. In Pythia-160M it moves from 0.45 to 0.58. As a result the coherence $\kappa$ falls during training, from 0.67 to 0.36 in Pythia-70M and from 0.52 to 0.22 in Pythia-160M.

\paragraph{Reading.} Training produces layers that are individually more sensitive to perturbations, while the expanding directions of consecutive layers become less aligned, so the forward pass as a whole stretches perturbations only moderately. The depth-averaged exponent settles early, while the per-layer exponents keep changing. This is consistent with the observation that the normalized singular value spectra of the weights stabilize early in pretraining \citep{zhang2026stability}, now seen through the Jacobians of the forward map. These are measurements on two small models with short contexts and a small prompt set. Whether the same pattern holds at scale, and whether $\kappa$ predicts robustness to perturbations or the success of model editing, is part of \op{1}.

\subsection{Experiment 2: a learning-rate bifurcation diagram (clock 2)}
\label{sec:exp2}

\paragraph{Question and setup.} How does the long-run behavior of gradient descent on a transformer change with the step size $\eta$? We train a one-block transformer (two heads, width 32) with full-batch gradient descent on next-token prediction of sequences from a fixed random Markov chain, for 56 step sizes between $0.02$ and $4$. Within one seed every step size starts from the same initialization and sees the same data, and we repeat the sweep for three seeds that change both. Full batches remove minibatch noise, so any irregularity comes from the map \eqref{eq:clock2} itself. For each run we record the last 400 losses and the sharpness $\lambda_{\max}$ at the end.

\begin{figure}[t]
\centering
\includegraphics[width=\textwidth]{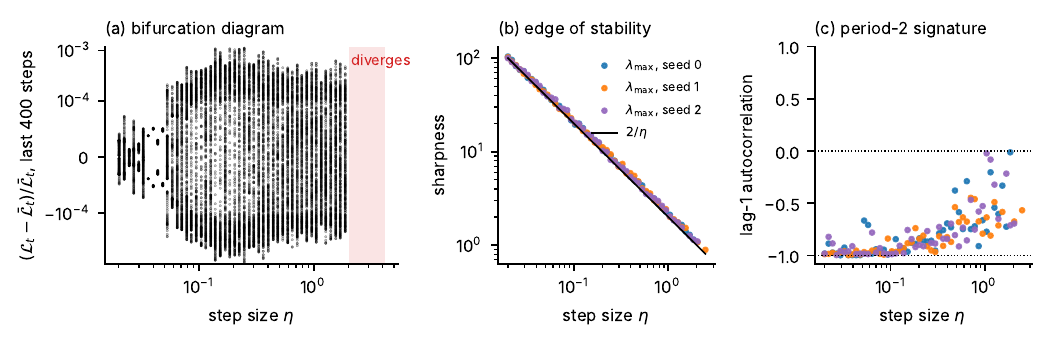}
\caption{Experiment 2: full-batch gradient descent on a one-block transformer. (a) Bifurcation diagram for seed 0: relative deviation of the last 400 losses from their running mean (9-step window, symmetric-log axis) against the step size. Two branches at small $\eta$ mark a period-2 orbit; the band at larger $\eta$ is irregular oscillation; the shaded region diverges. (b) Sharpness at the end of training against $2/\eta$ for three seeds. (c) Lag-1 autocorrelation of the detrended loss for the same runs; $-1$ is a clean period-2 orbit.}
\label{fig:exp2}
\end{figure}

\paragraph{Results.} Figure~\ref{fig:exp2} shows three regimes. In each seed, 48 of the 56 step sizes train. Across these 144 runs the final sharpness sits just above the stability threshold, with a median $\eta\lambda_{\max}/2$ of $1.07$ and a range of $[1.00,1.21]$ across two decades of step size (panel b). This is the edge-of-stability behavior of \citet{cohen2021gradient}, and in this model it holds over the entire trainable range we tested. None of the runs settles on a fixed point. Removing the slow downward trend of the loss shows what the iterate does instead (panels a and c). For $\eta<0.05$ the fluctuations alternate in sign at every step, a period-2 orbit riding on the slow descent, and the median lag-1 autocorrelation of the detrended loss over the ten smallest step sizes is $-0.99$, $-0.99$ and $-0.97$ in the three seeds. As $\eta$ grows the fluctuations become larger and less regular. The median lag-1 autocorrelation rises to between $-0.95$ and $-0.92$ for $0.05\le\eta<0.3$, between $-0.74$ and $-0.68$ for $0.3\le\eta<1$, and between $-0.66$ and $-0.32$ for $\eta\ge1$. The edge of trainability depends on the seed. In seed 0 every $\eta\ge2.04$ diverges, while in seeds 1 and 2 divergent and trainable step sizes interleave between $1.5$ and $2.5$, and in seed 1 the largest trainable step size (2.47) lies above the smallest divergent one (1.68).

\paragraph{Reading.} In dictionary terms, the trainable range of this model is a sequence from a period-2 orbit near the stability boundary, through increasingly irregular oscillation, to divergence. This is the picture that minimal models predict \citep{chen2022edge,zhu2022edgeofstability,kong2020stochasticity}, here measured on a transformer with attention and layer normalization. The interleaving of trainable and divergent step sizes near the edge is consistent with the fractal trainability boundary reported by \citet{sohldickstein2024boundary}, although a grid of 56 points cannot measure a fractal dimension. We do not claim a full period-doubling cascade, since with 400-step tails on top of a slowly drifting loss we cannot resolve periods above two.

\paragraph{Why the sharpness sits above $2/\eta$.} In 26 trainable runs we continued gradient descent and computed, at consecutive iterates, the top six Hessian eigenpairs, the curvature along the step and the finite-time Lyapunov spectrum of the gradient-descent map (Appendix~\ref{app:exp2}). Two to six eigenvalues exceed $2/\eta$ at every iterate (median four), and the sharpness sits 5\% to 14\% above the threshold, yet the curvature averaged along the step equals $2/\eta$ to within 0.2\% in every run, as the identity of \citet{litman2026origin} requires on average. The top eigenvectors at consecutive iterates are nearly orthogonal (median 86 degrees), so the top of the spectrum is a cluster whose members exchange order at every step, and the largest finite-time Lyapunov multiplier per two steps stays between 0.9998 and 1.019. The edge in this model is therefore a multi-mode regime with neutral or weakly chaotic dynamics, not a single stiff mode on a periodic orbit, and a warm-started power iteration, which averages over the alternating cluster, underestimates how far the sharpness sits above $2/\eta$ (\op{10}).

\subsection{Experiment 3: escape from repetition loops (clock 3)}
\label{sec:exp3}

\paragraph{Question and setup.} Section~\ref{sec:loops} predicts that escape from a repetition loop follows an Arrhenius law with the logit gap as barrier (Proposition~\ref{prop:arrhenius}), and that loops lock if the barrier grows from cycle to cycle (Proposition~\ref{prop:lock}). We test both on GPT-2 small \citep{radford2019language} and on the final checkpoint of Pythia-160M \citep{biderman2023pythia}. From each of 40 prompts we decode greedily for 320 tokens and keep the runs that end in a loop. Because the context along a loop is deterministic, the logits stored during the greedy run give the exact probability of staying in the loop at any temperature, as also used by \citet{chen2026repetition}, and we check these exact values against Monte Carlo sampling.

\begin{figure}[t]
\centering
\includegraphics[width=\textwidth]{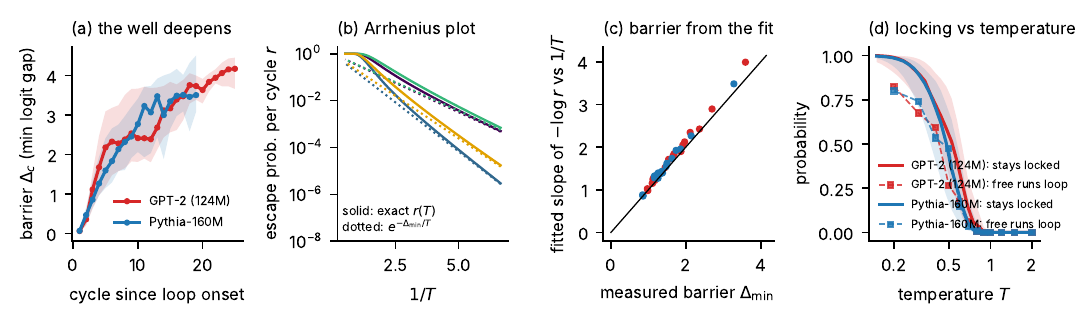}
\caption{Experiment 3: escape from repetition loops. (a) Barrier $\Delta_c$ (smallest logit gap in cycle $c$) against cycles since loop onset; line: median over loops, band: interquartile range; only cycles reached by at least 10 loops are shown. (b) Exact escape probability per cycle for four GPT-2 loops (solid) and the Arrhenius term $e^{-\Delta_{\min}/T}$ (dotted). (c) Barrier recovered from the slope of $\log r$ against $1/T$ at low temperature, against the measured $\Delta_{\min}$; line: identity. (d) Solid: median probability of staying in the loop until the end of the 320-token horizon, after two cycles of burn-in (band: interquartile range); the number of remaining cycles differs between loops because their onsets differ. Dashed: fraction of free generations of 200 tokens that end in a loop (120 per temperature).}
\label{fig:exp3}
\end{figure}

\paragraph{Results.} Greedy decoding ends in a loop for 37 of the 40 prompts in each model, with periods from 4 to 53 tokens (median 23) in GPT-2 and from 7 to 46 tokens (median 22) in Pythia-160M (Figure~\ref{fig:exp3}).

\paragraph{The barrier grows.} The median smallest logit gap $\Delta_c$ rises from 0.36 in the second cycle of a loop to 2.33 in the sixth and 3.63 in the twentieth in GPT-2, and from 0.47 to 1.83 and then 2.77 by the tenth cycle in Pythia-160M (panel a). The barrier in cycle 6 is larger than in cycle 2 for 36 of 36 GPT-2 loops and 35 of 37 Pythia-160M loops. This is the self-reinforcement reported by \citet{xu2022learning}, measured here as the depth of a well.

\paragraph{Escape follows the Arrhenius law.} For the first cycle after two cycles of burn-in, the escape probability per cycle approaches $e^{-\Delta_{\min}/T}$ at low temperature (panel b). Fitting $\log r$ against $1/T$ recovers the measured barrier with a correlation of 0.997 in both models and a median relative difference of 6.9\% (22 GPT-2 loops) and 5.7\% (21 Pythia-160M loops). Loops whose escape probability never falls below $10^{-2}$ on our temperature grid are excluded from the fit (panel c). The exact survival probabilities agree with Monte Carlo sampling. Across 24 comparisons per model (four loops, two temperatures, survival after 1, 2 and 4 cycles, 128 samples each), the largest absolute difference is 0.062 for GPT-2 and 0.068 for Pythia-160M, and every difference is within two binomial standard errors.

\paragraph{Loops lock below a fairly sharp temperature.} The median probability of staying in a loop until the end of the 320-token horizon is 0.89 at $T=0.31$ in both models, crosses 0.5 at $T\approx0.53$ (GPT-2) and $T\approx0.48$ (Pythia-160M), and falls to 0.06 and 0.015 at $T=0.75$ (panel d). Free sampling gives a consistent picture. With three samples of 200 tokens per prompt and temperature, 68\% (GPT-2) and 74\% (Pythia-160M) of generations end in a loop at $T=0.3$, 27\% and 48\% at $T=0.5$, at most 6\% at $T=0.7$, and none at $T\ge0.9$.

\paragraph{The barriers saturate, so locking depends on the horizon.} Fitted from cycle 2 on, the barrier of most loops follows a saturating law, $\Delta_c=\Delta_\infty-(\Delta_\infty-\Delta_2)\,e^{-(c-2)/c_\ast}$, better than a logarithmic or a linear one (21 of 36 GPT-2 loops and 26 of 37 Pythia-160M loops by the corrected Akaike criterion), with saturation levels of median 4.2 and 3.6 reached within 3 to 4 cycles. The second case of Proposition~\ref{prop:lock} therefore applies: no loop locks in an unbounded generation, and the locking temperature is a finite-horizon quantity. Recomputed from the exact escape probabilities for horizons of 1 to 1024 cycles, the median $T_{1/2}$ falls from 0.67 to 0.40 in GPT-2 and from 0.62 to 0.37 in Pythia-160M, and \eqref{eq:thalf} fits this decline with a largest deviation of 0.009 and 0.019. Appendix~\ref{app:exp3} gives the fits and the convergence of the hidden state onto a periodic orbit, which is what makes the barriers saturate (Figure~\ref{fig:exp3d}).

\begin{figure}[t]
\centering
\includegraphics[width=\textwidth]{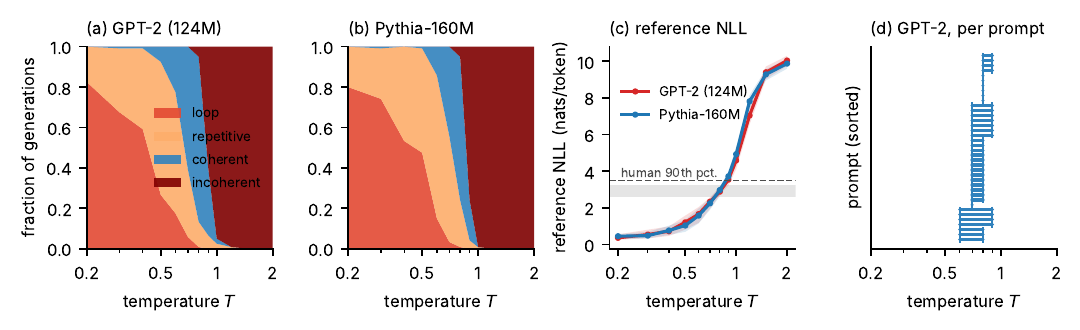}
\caption{Experiment 3, the coherent window. (a, b) Fraction of free generations (three samples of 200 tokens for each of 40 prompts) that end in a loop, are repetitive (repeated 4-grams above the 95th percentile of human text), are incoherent (reference score above the 90th percentile of human text) or are coherent, against temperature. (c) Median reference score in nats per token (band: interquartile range); gray band: interquartile range of human text; dashed: the incoherence threshold. (d) Coherent temperature range of each GPT-2 prompt, sorted; two prompts with an empty range are not shown.}
\label{fig:exp3b}
\end{figure}

\paragraph{Coherent text lives in a narrow window between two failures.} Raising $T$ ends the loops but eventually produces incoherent text, so we map both failures with the same free generations (Figure~\ref{fig:exp3b}). A generation counts as repetitive if its fraction of repeated 4-grams exceeds the 95th percentile of human continuations of 200 OpenWebText passages, and as incoherent if its mean negative log-likelihood under a separate reference model (Qwen2.5-0.5B) exceeds the 90th percentile of the human scores. Exact loops fall below one half at $T\approx0.43$ (GPT-2) and $T\approx0.45$ (Pythia-160M), loops and repetitive text together only at $T\approx0.67$ and $0.72$, and incoherent text rises above one half at $T\approx0.88$ and $0.86$. The coherent fraction peaks at $T=0.8$, at 82\% for GPT-2 and 71\% for Pythia-160M, and is below 25\% at $T=0.6$ and at $T=0.9$. Per prompt, the coherent range covers at most four of our twelve temperatures, typically two, and is empty for 2 of 40 prompts in GPT-2 and 5 of 40 in Pythia-160M. These proxies measure surprise and repetition, not meaning (Appendix~\ref{app:exp3}).

\paragraph{Larger models dig deeper wells.} We repeated the loop analysis on Pythia-70M, 410M and 1B, which share training data and order with Pythia-160M \citep{biderman2023pythia}. The saturated barrier rises with size (medians 3.0, 3.6, 4.5 and 4.8 from 70M to 1B), and so does the temperature needed to leave a loop: the median $T_{1/2}$ for a horizon of 16 cycles is 0.43, 0.48, 0.59 and 0.64. Over a 14-fold increase in parameters the locking temperature rises by about half, while the coherent window of the larger models is wider (Figure~\ref{fig:exp3e}).

\paragraph{Reasoning models and the recommended temperature.} Vendors of reasoning models prescribe a narrow sampling range: the DeepSeek-R1 model card asks for a temperature between 0.5 and 0.7 ``to prevent endless repetitions or incoherent outputs'' \citep{deepseekr1card,guo2025deepseekr1}, and the Qwen3 card recommends $T=0.6$ in thinking mode \citep{qwen3card,yang2025qwen3}. On the first 18 problems of GSM8K \citep{cobbe2021training}, Qwen3-0.6B in thinking mode with a budget of 1024 tokens reaches the answer in 0.28 of the traces at $T=0.2$, 0.56 at 0.4, 0.61 at 0.6, and 0.22 at both 1.0 and 1.3 (Figure~\ref{fig:exp3f}; Fisher exact test, p-value 0.09 for 0.6 against 0.2 and 0.04 against 1.0). The window is there, centered on the vendor's value, but its lower edge is a different failure: no trace loops exactly, and at $T=0.2$ the model keeps returning to the same step with small rephrasings until 13 of 18 traces run out of budget. This is a recurrent orbit without a period, for which recurrence quantification \citep{eckmann1987recurrence,marwan2007recurrence} is the natural tool (\op{5}). Without thinking the model never loops, so the repetitive edge appears only in long traces. A single 0.6B model with a short budget tests the shape of the window, not the vendors' exact numbers.

\paragraph{Reading.} The escape problem on clock~3 has the structure of Kramers' problem, with two differences that matter. The noise acts on a discrete choice through a Gibbs distribution, so the Arrhenius form follows directly from the softmax (Proposition~\ref{prop:arrhenius}) and needs no diffusion approximation. In addition, the barrier is a function of the residence time, so a loop becomes harder to leave the longer the model repeats it. Panel (a) also shows that the deepening is not uniform: barriers often rise quickly over the first five cycles, pause, and rise again. Escape from a well whose depth grows with the time already spent in it is a non-stationary escape problem. Its mechanical analog is a particle in a well that deepens as the particle settles. How the deepening scales beyond 1B parameters is open (\op{5}).

\FloatBarrier
\section{Open problems}
\label{sec:open}

The experiments answered some questions and opened others. Each problem below comes from an empty or weak cell of Table~\ref{tab:dictionary}, from a result of Section~\ref{sec:experiments}, or from a place where two clocks meet, and each is stated with a first test. Problems \op{9} and \op{10} are conjectures that come from the mechanics of nonlinear oscillators; we include them because they make predictions that can be checked, and we do not claim evidence for them yet.

\begin{enumerate}[label=\textbf{OP\,\arabic*.}, leftmargin=*, itemsep=3pt]
\item \textbf{Coupling of clocks 1 and 2.} Which features of the Lyapunov profile of the forward pass settle before the loss does? In two small models the whole-stack exponent settles within 3\% of training while the per-layer exponents keep growing (Experiment~1). The next test is the full Jacobian spectrum of every layer across the public checkpoints of a model family above a billion parameters.

\item \textbf{Periodic orbits in depth.} Do models that reuse a block many times, such as looped and recurrent-depth models \citep{prairie2026parcae}, have periodic or quasi-periodic orbits in the residual stream, and do these orbits carry computation? Return maps of hidden states over many loop iterations would show them directly.

\item \textbf{The full Lyapunov spectrum of a language model.} How many directions expand along a forward pass and along a generation, and what is the resulting attractor dimension? QR-based algorithms \citep{engelken2020lyapunov,vogt2020lyapunov} make this feasible for models of GPT-2 size, and the exponent of clock~3 would turn exposure bias into a measured quantity.

\item \textbf{The type of a loss spike.} Are the loss spikes of pretraining crossings of the stability boundary, period-doubling cascades, catapults \citep{lewkowycz2020large} or escapes from a basin? Each type leaves a different signature in the curvature and in the update direction just before the spike, so open training logs and checkpoints can be classified by these signatures.

\item \textbf{Escape laws across samplers and scales.} How do nucleus sampling and repetition penalties change the barrier of Proposition~\ref{prop:arrhenius}, and do the saturated barrier and the locking temperature keep growing with model size beyond 1B parameters? Reasoning models add the recurrent orbits of Experiment~3, which have no period: do they have an escape law of their own, and do recurrence plots of the trace predict when a model will run out of budget? A harder version asks whether the onset of a loop and the coherent window of a prompt can be read from the prompt and the weights before any text is generated.

\item \textbf{Fractal trainability of transformers.} Does the fractal boundary between convergent and divergent hyperparameters \citep{sohldickstein2024boundary}, reproduced for a small network in Figure~\ref{fig:basins}, persist for transformers trained with Adam, and how does its dimension depend on depth? Fine grids over learning rate and warmup length give a first estimate.

\item \textbf{Metastability in generation.} Is the length of a coherent stretch of generated text before drift a metastable residence time, exponentially long in some barrier, like the cluster states in depth \citep{geshkovski2024dynamic}? Survival curves of coherence against temperature would test this.

\item \textbf{Koopman modes of a forward pass.} Can a linear operator fitted to residual-stream trajectories predict the effect of an intervention several layers ahead, and do its eigenfunctions align with known features? Dynamic mode decomposition on residual streams, compared with the tuned lens \citep{belrose2023eliciting}, is a direct test.

\item \textbf{Targeted energy transfer (conjecture).} In coupled mechanical oscillators, a nonlinear energy sink absorbs vibration energy from a primary system in one direction and irreversibly \citep{vakakis2009nonlinear,farid2016tuned}. We conjecture a similar one-way transfer of signal energy between spectral modes of the residual stream, with massive activations and attention sinks \citep{sun2024massive,xiao2023efficient} in the role of the sink. Tracking the energy of each singular mode of the layer Jacobians across depth and training would confirm or refute it.

\item \textbf{Escape under internal forcing (conjecture).} At the edge of stability the iterate oscillates along stiff directions, and the other directions see this oscillation as periodic forcing. A periodically forced particle escapes from a potential well when the forcing amplitude crosses a threshold that depends on the forcing frequency \citep{farid2019escape,farid2022escape}, and we conjecture that some loss spikes are escapes of this kind. Gradient descent on a quartic well coupled to an oscillating direction is the simplest test. Experiment~2 adds a related question: self-stabilization \citep{damian2022selfstabilization} explains a single mode at the edge, but our transformer has several, which exchange order at every step, and their averaged dynamics is open.

\item \textbf{Early warning signals.} Near a bifurcation, recovery from perturbations slows down and fluctuations grow. Do grokking, induction-head formation and loss spikes show rising autocorrelation and variance of the loss or of the gradient norm before they happen? Runs with a known transition step \citep{power2022grokking,olsson2022incontext} allow a direct check.

\item \textbf{Control of clock 3.} Can a controller built from the escape analysis keep generation coherent at a lower cost than repetition penalties, with Mirostat \citep{basu2020mirostat} and truncation sampling \citep{hewitt2022truncation} as baselines? Three first tests on GPT-2 (Appendix~\ref{app:exp3}) fail in an instructive way. Heating the sampler when a loop is detected cuts looping generations at $T=0.3$ from 68\% to 6\%, but 94\% of the outputs remain repetitive. Rolling the text back to the first token of a loop and resampling there removes all exact loops at $T=0.5$, but 91\% of the outputs stay repetitive. A feedback controller on recurrence and surprisal settles near the center of the coherent window from any start, but its share of coherent texts (21\% to 83\%) does not beat the 82\% of the best fixed temperature. All three act on the noise, while the trap is stored in the context. Controllers that act on the state, by editing the cache or by adding a steering vector to the residual stream \citep{turner2023steering,zou2023representation}, are the natural next step: in mechanical terms, a control force on the state instead of a change of the noise intensity.
\end{enumerate}

\section{Conclusion}
\label{sec:conclusion}

A transformer language model is iterated in three ways: over layers, over optimizer steps and over generated tokens. Each iteration has been studied with tools from dynamical systems, mostly by different communities and under different names. We gave the three iterations one formal frame, reviewed the results on each, and summarized them in a dictionary that records which concepts have been proved, measured or only conjectured on which clock. In experiments on open models we found that training makes each layer of a Pythia model more expansive while the coherence of the whole forward pass falls, and that gradient descent on a small transformer stays at a multi-mode edge of stability across two decades of step size. Repetition loops in GPT-2 and Pythia behave as wells whose depth grows while the model sits in them and then saturates. Whether a loop locks therefore depends on the length of the generation and on the size of the model, and coherent text lies in a narrow temperature window between repetition and incoherence. A small reasoning model has the same window, centered on its recommended temperature, and its low-temperature failure is recurrent rethinking, with no periodic loop.

A bridge between two fields can describe, explain and solve. This paper describes the three clocks in one vocabulary, shows the correspondences side by side, derives the reduced-order models behind them, and explains several behaviors that the machine learning literature had measured but not accounted for, from the Arrhenius escape from loops to the narrow window of coherent text. It does not yet solve a problem of language models that was unsolved before: our controllers for repetition find the right temperature but do not beat it. We think the third level is within reach, because many problems of language models have counterparts that dynamics has already solved, and the open problems of Section~\ref{sec:open} point to where.

Many cells of the dictionary are empty, and several of the open problems sit where two clocks meet, such as the effect of training on the depth dynamics, the feedback of generated text into training, and the prediction of training instabilities with the tools used for escape problems in mechanics. Each of these questions has a counterpart that nonlinear dynamics has studied for decades on pendulums, ships and oscillator networks, and each can be tested with the weights and outputs of open models on a laptop. We hope that the shared vocabulary of this paper helps researchers from both communities recognize the same phenomenon when they meet it, and solve it together.

A further coupling lies outside the scope of this paper. When a model receives feedback from tools or from an environment, as in agents and world models, clock 3 is no longer a closed system but a feedback loop, and the external signal can break a repetition loop or couple with it. Feedback with a delay is known from mechanics to destabilize systems that are stable without it, and whether this happens in agent loops is open. In a world model the latent state has a fixed dimension, which makes the physical analogy closer than it is for a context that grows at every step.

\subsubsection*{Code and data}
Code to reproduce every figure and number, the raw results, and a maintained list of the papers in this review are available \codeurl. All experiments run on a CPU.

\subsubsection*{Broader impact statement}
This is a review with small-scale experiments on public models, and we do not foresee direct negative consequences. A better account of repetition loops and of training instabilities could reduce wasted compute in pretraining and make generation more reliable.

\subsubsection*{Acknowledgments}
We thank Amotz Thee for suggesting the extension to agents and world models.

\subsubsection*{Use of AI tools}
\label{sec:ai}
An AI assistant (Claude, Anthropic) was used to search and verify the literature, draft and edit text, and write and run experiment code under the author's direction. All references were checked against arXiv or Crossref records, all numbers come from the released code, and the author is responsible for every claim in the paper.

{\small\setlength{\bibsep}{2pt plus 0.5pt}
\bibliography{refs}}
\bibliographystyle{tmlr}

\appendix
\section{Glossary for readers from either field}
\label{app:glossary}

The first list defines terms of machine learning for readers from nonlinear dynamics, and the second defines terms of nonlinear dynamics for readers from machine learning. Each definition is the minimum needed to follow the paper.

\subsection*{Terms from machine learning}
\begin{description}[leftmargin=0pt,itemsep=1pt,font=\normalfont\bfseries]
\item[Token, vocabulary.] The unit of text a model reads and writes, usually a word or a word fragment such as ``ing''. GPT-2's vocabulary $\mathcal V$ holds about 50{,}000 of them, so each step of clock~3 picks one of 50{,}000 states.
\item[Context.] Everything the model has seen so far, prompt plus generated text, on which its next prediction is conditioned.
\item[Embedding, hidden state, residual stream.] The state of clock~1. Token $i$ starts as a vector in $\mathbb{R}^d$ (its embedding) and after block $\ell$ is the hidden state $x_i^\ell$. Across depth this vector is called the residual stream, since each block reads it and adds its own update.
\item[Transformer block.] One step of clock~1. Attention first mixes information across tokens, then a multilayer perceptron (MLP) acts on each token alone, and each of the two outputs is added back to the hidden state through a skip (residual) connection.
\item[Attention.] The coupling between tokens in clock~1. Each token takes an average of the hidden states of earlier tokens, weighted by a softmax of inner products between learned projections (queries and keys).
\item[Layer normalization.] Each hidden state is shifted and scaled to zero mean and unit variance before entering a sublayer, so the tokens stay close to a sphere.
\item[Logits, softmax.] One real score $z_y$ per vocabulary token, read off the last hidden state; in Section~\ref{sec:loops} minus the logit plays the role of an energy. Sampling uses $e^{z_y}/\sum_{y'}e^{z_{y'}}$ (the softmax), which is the Boltzmann distribution at $T=1$.
\item[Sampling temperature.] The number $T$ that divides the logits before the softmax, and the noise intensity of clock~3. As $T\to0$ the sampler always takes the top token (greedy decoding); at large $T$ every token becomes nearly equally likely.
\item[Top-k and nucleus (top-p) sampling.] Truncation rules that sample only from the $k$ most likely tokens, or from the smallest set whose probabilities sum to $p$.
\item[Autoregressive generation.] Writing text one token at a time, appending each sampled token to the context before predicting the next.
\item[Loss, cross-entropy.] The potential $\mathcal L$ that clock~2 descends. For next-token prediction it averages $-\log p$ over the training text, with $p$ the probability the model gave to the token that actually came next.
\item[Surprisal, perplexity.] The surprisal of a token is $-\log p$, in nats in this paper; perplexity is $e$ raised to the mean surprisal.
\item[Minibatch, SGD, learning rate.] Stochastic gradient descent (SGD) estimates the gradient on a random subset of the data (a minibatch) and steps against it with step size $\eta$, the learning rate.
\item[Adam, preconditioner.] Adam rescales each coordinate of the gradient by running estimates of its first and second moments, which acts as a diagonal preconditioner.
\item[Pretraining, checkpoint.] Training a model on a large text corpus to predict the next token; a checkpoint is a saved copy of the weights at one training step.
\item[Teacher forcing, exposure bias.] During training the model always conditions on true text; during generation it conditions on its own samples, and the resulting mismatch is called exposure bias.
\item[Sharpness.] The largest eigenvalue $\lambda_{\max}$ of the Hessian of the loss.
\item[Loss spike, grokking, induction head.] A sudden, temporary rise of the training loss; a late jump in test accuracy long after the training loss has saturated; an attention circuit that copies a token that followed an earlier occurrence of the current token, which forms abruptly during training.
\item[In-context learning.] Learning a task from examples given in the prompt, without changing the weights.
\item[Interpretability, tuned lens.] The study of what the internal states of a network represent; the tuned lens decodes a next-token prediction from an intermediate layer.
\end{description}

\subsection*{Terms from nonlinear dynamics}
\begin{description}[leftmargin=0pt,itemsep=1pt,font=\normalfont\bfseries]
\item[Map, flow.] A discrete-time system $u_{t+1}=F(u_t)$; a flow is its continuous-time counterpart $\dot u=f(u)$.
\item[Fixed point, linear stability.] A state with $F(u^\ast)=u^\ast$; it is linearly stable when every eigenvalue of the Jacobian $\mathrm{D}F(u^\ast)$ has modulus below one, so small perturbations decay.
\item[Periodic orbit, limit cycle.] A sequence of states that repeats with period $k$; an isolated attracting periodic orbit is a limit cycle.
\item[Attractor, basin of attraction.] A set that nearby trajectories approach; its basin is the set of initial states whose trajectories end on it.
\item[Bifurcation.] A qualitative change of the long-term behavior as a parameter crosses a critical value; in a period-doubling (flip) bifurcation an eigenvalue crosses $-1$ and a period-2 orbit appears, and in a saddle-node bifurcation two fixed points collide and vanish.
\item[Ghost.] The slow passage of trajectories through the region where a fixed point existed just before a saddle-node bifurcation.
\item[Chaos, Lyapunov exponent.] Chaos is bounded, aperiodic motion with sensitive dependence on initial conditions; the largest Lyapunov exponent is the mean exponential rate at which nearby trajectories separate, and a finite-time exponent measures it over a finite number of steps.
\item[Fractal boundary.] A boundary between basins with structure on every scale, so that arbitrarily small changes of the initial state can change the outcome.
\item[Metastable state.] A state that is not an attractor but holds trajectories for a very long time.
\item[Duffing oscillator.] A mass on a spring with a cubic term in its restoring force; with a negative linear stiffness its potential has two wells.
\item[Kramers escape, Arrhenius law.] The mean time for a noisy particle to leave a potential well of depth $\Delta U$ grows as $e^{\Delta U/D}$ with the noise intensity $D$; for chemical reactions this is the Arrhenius law with temperature in place of $D$.
\item[Kuramoto model, order parameter.] A model of oscillators that pull each other's phases together; the order parameter, the length of the mean of their phase vectors, measures how synchronized they are, from 0 to 1.
\item[Slow-fast system, multiple scales.] A system whose variables evolve on separated time scales; the method of multiple scales writes the solution as a function of a fast time and of slower times that differ by powers of a small parameter.
\item[Reduced-order model.] A low-dimensional model that keeps the term of the governing equations responsible for a behavior, such as an equivalent oscillator for a sloshing tank.
\item[Time history, power spectrum, phase portrait, recurrence plot.] Four standard pictures of a measured signal: the signal against time, its power at each frequency, the trajectory in the plane of two state variables, and the matrix of pairs of times at which the state returns close to itself.
\item[Koopman operator.] The linear operator that advances functions of the state (observables) along a nonlinear map.
\end{description}

\section{Notation}
\label{app:notation}

Each symbol has one meaning throughout the paper. The indices $i$ and $j$ count tokens (and $i$ the steps within a loop); their range is given where they appear.

{\small
\setlength{\LTpre}{2pt}\setlength{\LTpost}{2pt}
\begin{longtable}{@{}p{0.2\linewidth}p{0.76\linewidth}@{}}
\toprule
Symbol & Meaning \\
\midrule
\endhead
\bottomrule
\endfoot
\multicolumn{2}{@{}l}{\emph{Language model and clock 1 (layer time)}}\\
$L$ & number of residual blocks (depth) \\
$\ell$ & block index; the layer time of clock~1 \\
$d$ & model width \\
$\mathcal V$, $|\mathcal V|$ & vocabulary; its size \\
$x^{\ell}_i$, $x^{\ell}$ & hidden state (residual stream) of token $i$ after block $\ell$; all tokens together \\
$f^{\ell}_\theta$ & update added by block $\ell$ (attention and MLP) \\
$F$, $F_\ell$ & a generic map; the map of block $\ell$ \\
$\mathrm{D}F$ & Jacobian of $F$ \\
$z$, $z_y$ & logits; logit of token $y$ \\
$W_U$; $W_Q,W_K,W_V$ & unembedding matrix; query, key and value matrices \\
$\beta$ & inverse temperature of attention \\
$Z_i$ & normalizer of the attention weights of token $i$ \\
$\Pi_x$ & projection onto the tangent space of the unit sphere at $x$ \\
$\phi$ & activation function of a fully connected network (Section~\ref{sec:edgechaos}) \\
$\sigma_w^2$, $\sigma_b^2$ & variances of the weights and biases at initialization \\
$Q^\ast$, $G$ & fixed point of the length map; a standard Gaussian random variable \\
$\chi_1$ & factor by which one layer multiplies the distance between nearby inputs \\
\midrule
\multicolumn{2}{@{}l}{\emph{Clock 2 (training time)}}\\
$t$ & optimizer step; the training time of clock~2 \\
$\varepsilon$; $\tau_0,\tau_1,\tau_2$ & small parameter of a multiple-scales expansion ($\varepsilon=\eta$ in Section~\ref{sec:maps}); fast and slow times $\tau_k=\varepsilon^k t$ \\
$\theta$, $\theta_t$, $\theta^\ast$ & all trainable parameters; at step $t$; at a minimum \\
$\mathcal L$, $\mathcal L(\theta;\mathcal B_t)$ & training loss; loss on the minibatch $\mathcal B_t$ \\
$\mathcal B_t$ & minibatch of training sequences used at step $t$; the sequence $\mathcal B_0,\mathcal B_1,\dots$ is the data stream \\
$B$ & batch size \\
$\eta$, $\eta_t$ & learning rate (step size); at step $t$ \\
$P_t$ & preconditioner (identity for gradient descent) \\
$g_t$ & gradient at step $t$ \\
$H$ & Hessian of the loss \\
$\lambda$, $\lambda_{\max}$ & eigenvalue of the Hessian; the largest one (sharpness) \\
$\lambda_t$ & sharpness at step $t$ in the self-stabilization map \eqref{eq:selfstab} \\
$v$ & top eigenvector of the Hessian \\
$\mathcal L_v$ & loss restricted to the line through $\theta^\ast$ along $v$ \\
$\alpha$ & rate of progressive sharpening in \eqref{eq:selfstab} \\
$C$, $\Sigma$ & covariance of per-example gradients; stationary covariance of SGD near a minimum \\
$\mathcal W_t$ & Wiener process \\
$w_1,w_2$ & the two weights of the catapult model (Figure~\ref{fig:catapult}) \\
\midrule
\multicolumn{2}{@{}l}{\emph{Clock 3 (generation time)}}\\
$n$ & number of tokens in the context; the generation time of clock~3 \\
$y_n$, $s_n$ & token at position $n$, an element of $\mathcal V$; context $s_n=(y_1,\dots,y_n)\in\mathcal V^n$, whose length grows by one per step of clock~3 \\
$y$ & a token of the vocabulary \\
$T$, $T_0$ & sampling temperature; base temperature of a controller \\
$\pi_T(\cdot\mid s_n)$ & next-token distribution at temperature $T$, a probability vector with $|\mathcal V|$ entries \\
$E$, $E_y$, $k_B$ & energy; energy of token $y$, $E_y=-z_y$; Boltzmann constant, set to 1 \\
$k$ & period of a loop in tokens; period of an orbit \\
$a_i$ & $i$-th token of a loop \\
$p_i(T)$ & probability that the sampler stays on the loop at step $i$ \\
$S_i(T)$ & sum of the Boltzmann factors of all exits at step $i$ \\
$\Delta_i^{y}$, $\Delta_i$ & logit gap between the loop token $a_i$ and token $y$; smallest gap at step $i$ \\
$\Delta$, $\Delta_{\min}$ & barrier of a loop: its smallest logit gap \\
$m$, $m_i$ & number of exits that attain the smallest gap (of the cycle; at step $i$) \\
$c$ & cycle index of a loop \\
$\Delta_c$ & smallest logit gap during cycle $c$ \\
$\Delta_1$, $\Delta_2$, $\Delta_\infty$ & barrier in cycles 1 and 2; its saturation level \\
$b$, $\gamma$ & growth rate of the barrier under linear and logarithmic growth \\
$c_\ast$ & number of cycles over which the barrier saturates \\
$r(T)$, $r_c(T)$ & escape probability per cycle; during cycle $c$ \\
$r_\ast$ & target escape probability of the Arrhenius-guided controller \\
$\nu$ & number of escape attempts per cycle in the reduced model \eqref{eq:reduced} \\
$N$ & horizon: number of cycles (or forcing periods) observed \\
$T_{1/2}(N)$ & temperature at which a loop survives $N$ cycles with probability one half \\
$T_{\mathrm{lock}}$ & temperature below which a loop is never left with positive probability \\
$\mathrm{dist}_c$ & distance between the hidden state and its value one period earlier, in cycle $c$ \\
$\mathrm{rep}$, $\mathcal I$ & fraction of repeated 4-grams in a window; mean surprisal of the window \\
\midrule
\multicolumn{2}{@{}l}{\emph{Physical and reduced models}}\\
$\tau$, $\bar\tau$ & physical time; mean time to escape from a well \\
$q$, $\dot q$ & displacement of an oscillator or particle (or along $v$); velocity \\
$M$, $\zeta$ & mass; friction coefficient \\
$k_1$, $k_3$ & linear and cubic stiffness \\
$A$, $\Omega$ & amplitude and frequency of harmonic forcing \\
$\xi$, $D$ & Gaussian white noise; its intensity \\
$U$, $\Delta U$ & potential energy; height of the barrier \\
$\omega_{\mathrm{w}}$, $\omega_{\mathrm{b}}$ & frequencies set by the curvature at the bottom of the well and at the top of the barrier \\
$\psi_i$, $\omega_i$ & phase and natural frequency of oscillator $i$ \\
$J$ & coupling strength of the Kuramoto model \\
$\rho$, $\rho_\ell$ & order parameter of synchrony; of the residual stream entering block $\ell$ \\
$u$, $\mu$ & state of a generic map; parameter of the logistic map \\
\midrule
\multicolumn{2}{@{}l}{\emph{Dynamical-systems tools}}\\
$\Lambda_K$ & finite-time Lyapunov exponent over $K$ steps \eqref{eq:ftle} \\
$\Lambda$, $\tilde\Lambda_\ell$ & exponent of the whole stack; relative exponent of block $\ell$ (Section~\ref{sec:exp1}) \\
$\sigma_{\max}$, $\sigma_\ell$ & largest singular value; largest singular value of the Jacobian of block $\ell$ \\
$\kappa$ & coherence of stretching between blocks (Section~\ref{sec:exp1}) \\
$\delta$ & size of an initial perturbation \\
$R_{ij}$, $\epsilon_R$ & recurrence matrix; recurrence threshold \\
$\mathcal H_s$, DET & normalized spectral entropy; determinism of a recurrence plot \\
$h$ & an observable acted on by the Koopman operator \\
$I$ & identity matrix \\
$O$, $o$, $\Theta$ & asymptotic orders \\
$\mathcal N(0,\cdot)$ & Gaussian distribution with mean zero and the stated variance \\
\end{longtable}
}

\section{Proofs}
\label{app:proofs}

\paragraph{Proof of Proposition~\ref{prop:arrhenius}.}
Since $1-\prod_i p_i\le\sum_i (1-p_i)$ for $p_i\in[0,1]$, and $1-p_i=S_i/(1+S_i)\le S_i$, the bound follows. As $T\to0$, each $S_i(T)$ is dominated by its smallest gaps, $S_i(T)=m_i e^{-\Delta_i/T}(1+o(1))$ with $\Delta_i=\min_y\Delta_i^y$, and $1-\prod_i p_i=\sum_i S_i+O(\max_i S_i^2)$. The terms with $\Delta_i=\Delta_{\min}$ dominate the sum.
 $\square$

\paragraph{Proof of Proposition~\ref{prop:lock}.}
By the bound in Proposition~\ref{prop:arrhenius}, $r_c\le k(|\mathcal V|-1)e^{-\Delta_c/T}$, which is summable under linear growth, and $\prod_c(1-r_c)\ge1-\sum_c r_c$. If the logits converge, then $r_c$ tends to a positive limit and the second statement follows from the first part of Proposition~\ref{prop:arrhenius}.
 $\square$

\paragraph{Proof of Proposition~\ref{prop:tlock}.}
At the step of cycle $c$ that attains $\Delta_c$, $1-p_i=S_i/(1+S_i)\ge\tfrac12\min(1,e^{-\Delta_c/T})$, so $r_c\ge\tfrac12\min(1,e^{-\Delta_c/T})$, and Proposition~\ref{prop:arrhenius} gives $r_c\le k(|\mathcal V|-1)e^{-\Delta_c/T}$. For large $c$ both bounds are constant multiples of $c^{-\gamma/T}$, so $\sum_c r_c$ converges exactly when $\gamma/T>1$, and $\prod_c(1-r_c)>0$ exactly when the sum converges. For $T>\gamma$, $r_c\to0$, so $-\log\prod_{c\le N}(1-r_c)$ is within constant factors of $\sum_{c\le N}c^{-\gamma/T}=\Theta(N^{1-\gamma/T})$.
 $\square$

\section{Experimental details}
\label{app:details}

All experiments run on a CPU with two cores and 7\,GB of memory, using PyTorch and Hugging Face Transformers with the public checkpoints of GPT-2 small \citep{radford2019language} and Pythia \citep{biderman2023pythia}. The repository contains one script per experiment, the exact command lines, the raw JSON outputs and the plotting scripts. The 40 prompts used in Experiments~1 and~3 are short neutral sentence openings written for this study (for example, ``The weather today is cold and cloudy, with a chance of''), and the full list is in the repository.

\subsection{Experiment 1: finite-time Lyapunov exponents}
\label{app:exp1}

For each prompt we take the first $n$ tokens (prompts shorter than $n$ are repeated) and run the model once to record the residual stream $x^0,\dots,x^L$ before the final layer normalization. We then call each block of the Hugging Face implementation directly as a function $F_\ell:\mathbb{R}^{n\times d}\to\mathbb{R}^{n\times d}$, with rotary embeddings and an explicit causal mask, and check that chaining these calls reproduces the model's hidden states (the maximum relative deviation was zero in float32 for every checkpoint). The largest singular value of the Jacobian $\mathrm{D}F_\ell(x^\ell)$ is estimated by power iteration on $\mathrm{D}F_\ell^{\top}\mathrm{D}F_\ell$, using forward-mode (JVP) and reverse-mode (VJP) products from \texttt{torch.func}. The same procedure applied to the composition $F_{L-1}\circ\cdots\circ F_0$ gives the depth-averaged exponent $\frac{1}{L}\log\sigma_{\max}(\mathrm{D}F_{0\to L})$. We use eight prompts, contexts of 16 tokens and 10 power iterations per estimate, for eleven Pythia-70M checkpoints (steps 0, 128, 512, 1k, 2k, 4k, 8k, 16k, 32k, 64k and 143k) and six Pythia-160M checkpoints (steps 0, 512, 2k, 8k, 32k and 143k).

\subsection{Experiment 2: bifurcation diagram}
\label{app:exp2}

\paragraph{Detail: Why the sharpness sits above $2/\eta$.} In 26 trainable runs (three seeds, eight or nine step sizes each) we continued gradient descent for 200 to 400 steps and computed, at consecutive iterates, the top six Hessian eigenpairs by Lanczos iteration, the curvature along the step, a finite-time Lyapunov spectrum of the gradient-descent map and the leading multipliers of the two-step Jacobian by Arnoldi iteration (setup below). The sharpness at the iterates sits 5\% to 14\% above the threshold ($\eta\lambda_{\max}/2$ between 1.05 and 1.14, median 1.11), and two to six eigenvalues exceed $2/\eta$ at every iterate (median four). The curvature of the Hessian averaged over the segment between consecutive iterates, taken along the step, equals $2/\eta$ to within 0.2\% in every run, as the telescoping identity of \citet{litman2026origin} requires on average. The stiff directions do not stay in place. The top eigenvectors at consecutive iterates are 74 to 88 degrees apart (median 86), so the top of the spectrum is a cluster whose members exchange order at every step, and the step is only partly aligned with the top eigenvector ($|\cos|$ between 0.08 and 0.42). Locally the two-step map expands, with a largest multiplier between 1.0002 and 1.126 that is often a slow complex pair near $+1$, yet along the trajectory the largest finite-time Lyapunov multiplier per two steps lies between 0.9998 and 1.019 while the loss stays bounded. In this model the edge is therefore a multi-mode regime with neutral or weakly chaotic dynamics, not a single stiff mode on a periodic orbit. A warm-started power iteration, which we used in a first pass, gives ratios of only 1.02 to 1.07, because it averages over the alternating cluster, so reported values of how far the sharpness sits above $2/\eta$ depend on the eigensolver whenever the top of the spectrum is degenerate. Why the oscillation stays bounded with several modes above the threshold is open (\op{10}).

\emph{Data.} A first-order Markov chain over $|\mathcal V|=12$ symbols, with transition rows drawn from a Dirichlet distribution with concentration $0.3$ (seed 0), generates 64 sequences of length 13, and the model predicts the next symbol at each of the 12 positions. \emph{Model.} One pre-LN transformer block with causal attention (two heads, width 32, MLP width 64), learned positional embeddings and a linear readout, in float64. Attention is written out explicitly so that Hessian-vector products work on a CPU. \emph{Training.} Full-batch gradient descent on the mean cross-entropy, from the same initialization for every step size within a seed. Seeds 0, 1 and 2 change both the Markov chain and the initialization. Each run lasts $\max(3000,\,600/\eta)$ steps, capped at 30{,}000, so that all runs cover a comparable amount of gradient-flow time. A run is marked as diverged if the loss exceeds 50 or becomes non-finite. \emph{Measurements.} The last 400 losses of each run, and the top Hessian eigenvalue at the final iterate from 40 steps of power iteration on Hessian-vector products. The step sizes are 56 values spaced geometrically between $0.02$ and $4$.

\emph{Edge-of-stability geometry.} For each seed and for nine step sizes spaced geometrically between 0.05 and 1.6, we train as above and continue gradient descent. The curvature along the step is measured over 200 further steps as $\langle g_t-g_{t+1},g_t\rangle/(2\|g_t\|^2)$, which equals $\eta/2$ times the curvature of the segment-averaged Hessian along the step. Over a separate window of 400 steps we compute, every 50 steps at a pair of consecutive iterates, the top six Hessian eigenpairs by Lanczos iteration (\texttt{scipy} \texttt{eigsh} on Hessian-vector products, residuals near $10^{-13}$), their mutual angles and the alignment of the step with the top eigenvector; a finite-time Lyapunov spectrum of six vectors with QR reorthogonalization at every step; and, at step 200, the eight largest multipliers of $(I-\eta H_{t+1})(I-\eta H_t)$ by Arnoldi iteration. Power iteration is not suitable for this two-step map, since thousands of flat directions have multipliers near one. The largest step size (1.6) diverged in seed 0, leaving 26 runs. Per-run values are in \texttt{results/eos\_rotation/}.

\subsection{Experiment 3: escape from repetition loops}
\label{app:exp3}

\paragraph{Full results behind Section~\ref{sec:exp3}.} The following paragraphs give the complete numbers summarized in the main text.

\begin{figure}[t]
\centering
\includegraphics[width=\textwidth]{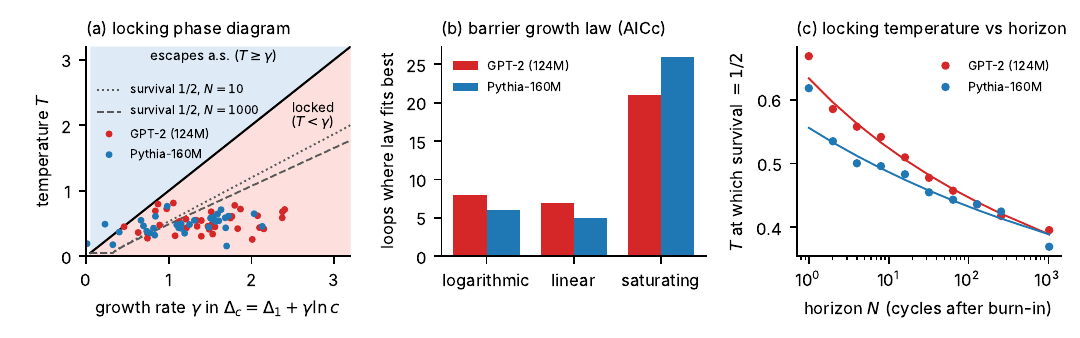}
\caption{Experiment 3, locking. (a) Phase diagram for the logarithmic barrier $\Delta_c=\Delta_1+\gamma\ln c$ (Proposition~\ref{prop:tlock}): loops lock with positive probability below $T=\gamma$ and escape almost surely above it. Dotted and dashed lines: temperature at which survival over $N=10$ and $N=1000$ cycles equals one half, for the single-gap rate with $m=1$ and the median fitted $\Delta_1$. Points: fitted $\gamma$ and exact finite-horizon $T_{1/2}$ of each loop. (b) Barrier growth law preferred by the corrected Akaike criterion. (c) Median exact $T_{1/2}$ against the horizon $N$ (points) and the fit of \eqref{eq:thalf} (lines).}
\label{fig:exp3d}
\end{figure}

\paragraph{Detail: The barriers saturate, so locking depends on the horizon.} For each loop with at least five cycles after the first, we fit $\Delta_c$ for $c\ge2$ with the logarithmic law of Proposition~\ref{prop:tlock}, a linear law and a saturating law $\Delta_c=\Delta_\infty-(\Delta_\infty-\Delta_2)\,e^{-(c-2)/c_\ast}$, and compare the fits by the corrected Akaike criterion (Figure~\ref{fig:exp3d}). The saturating law fits best for 21 of 36 GPT-2 loops and 26 of 37 Pythia-160M loops, the logarithmic law for 8 and 6, and the linear law for 7 and 5. The fitted saturation levels have medians 4.2 and 3.6, reached within a time constant of 3 to 4 cycles. The hidden state converges onto a periodic orbit at the same pace (end of this section), and since the logits are a linear function of the final hidden state, this convergence is what makes the barriers saturate. On these models the second case of Proposition~\ref{prop:lock} applies, and no loop locks in an unbounded generation. The logarithmic law would not explain the measured locking temperatures either. Its rate $\gamma$ has median 1.48 (GPT-2) and 1.15 (Pythia-160M), more than twice the measured $T_{1/2}$, and exceeds the loop's own $T_{1/2}$ in 35 of 35 and 32 of 35 loops (panel a). The locking temperature of Experiment~3 is therefore a finite-horizon quantity. We recompute it from the exact per-cycle escape probabilities for horizons of 1 to 1024 cycles after burn-in, repeating the last stored cycle beyond the end of the run, which matches the saturation. The median $T_{1/2}$ falls from 0.67 to 0.40 in GPT-2 and from 0.62 to 0.37 in Pythia-160M, and \eqref{eq:thalf} fits this decline with a largest deviation of 0.009 and 0.019 (panel c). The fitted constants are effective values ($\Delta_\infty$ of 7.0 and 9.0, $\ln m$ of 10.6 and 15.7). They exceed the measured gaps because at $T\approx0.5$ many competing tokens contribute to escape besides the nearest one, and the single-gap Arrhenius rate with $m=1$ overestimates $T_{1/2}$ by a median of 37\% (GPT-2) and 39\% (Pythia-160M), although it ranks the loops correctly (correlation 0.89 and 0.92).

\paragraph{Detail: Coherent text lives in a narrow window between two failures.} Raising $T$ ends the loops but eventually produces incoherent text, so we map both failures with the same free generations (Figure~\ref{fig:exp3b}). A generation counts as a loop by the test above. Otherwise it counts as repetitive if its fraction of repeated 4-grams exceeds the 95th percentile of human continuations of 200 OpenWebText passages (0.091), and as incoherent if its mean negative log-likelihood under a separate reference model (Qwen2.5-0.5B) exceeds the 90th percentile of the human scores (3.49 nats per token, against a human median of 2.92). The remaining generations count as coherent. Exact loops fall below one half at $T\approx0.43$ (GPT-2) and $T\approx0.45$ (Pythia-160M), but loops and repetitive text together fall below one half only at $T\approx0.67$ and $T\approx0.72$, and incoherent text rises above one half at $T\approx0.88$ and $T\approx0.86$. The coherent fraction peaks at $T=0.8$, at 82\% for GPT-2 and 71\% for Pythia-160M, and is below 25\% at $T=0.6$ and at $T=0.9$. Moving the incoherence threshold between the human median and the 95th percentile moves the peak between $T=0.7$ and $T=0.8$ and does not widen the window. For each prompt we also take the temperatures at which fewer than half of its samples loop and its median repetition and reference scores are within the human thresholds. This range covers at most four of our twelve temperatures, typically two, lies between 0.6 and 0.9, and is empty for 2 of 40 prompts in GPT-2 and 5 of 40 in Pythia-160M, so for these prompts no single temperature avoids both failures. At the upper edge the entropy of the sampling distribution is 2.3 to 3.1 nats, and at $T=2$ it reaches 10.3 nats, 95\% of $\log|\mathcal V|$, so sampling there is spread over almost the whole vocabulary. These proxies measure surprise and repetition, not meaning, and a reference model of 0.5B parameters is a coarse judge of coherence. Figure~\ref{fig:atlas} shows the three regimes of one prompt in the representations of Section~\ref{sec:regimes}: the loop is a periodic orbit with sharp spectral lines and long diagonal recurrence lines, and the determinism of the hidden-state trajectory falls from the loop to coherent and then to incoherent text.

\begin{figure}[t]
\centering
\includegraphics[width=\textwidth]{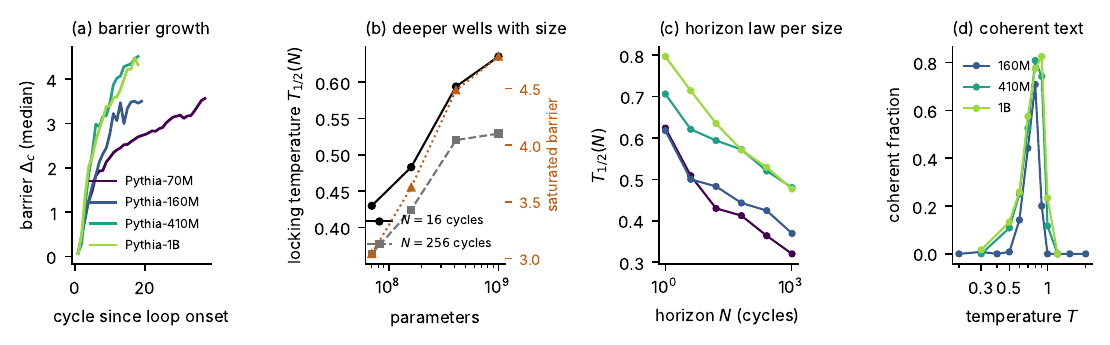}
\caption{Experiment 3 across model size, Pythia-70M to Pythia-1B (same data and training order). (a) Median barrier $\Delta_c$ against the cycle since loop onset (cycles reached by at least 10 loops). (b) Median locking temperature $T_{1/2}(N)$ for horizons of 16 and 256 cycles (left axis) and median saturated barrier $\Delta_\infty$ from the saturating fits (right axis) against the number of parameters. (c) $T_{1/2}(N)$ against the horizon for each model. (d) Fraction of coherent free generations against temperature for the three larger models.}
\label{fig:exp3e}
\end{figure}

\paragraph{Detail: Larger models dig deeper wells.} We repeated the loop analysis on Pythia-70M, 410M and 1B, which share training data, tokenizer and data order with Pythia-160M \citep{biderman2023pythia}, so model size is the only variable (Figure~\ref{fig:exp3e}). Greedy decoding loops for 33 to 37 of the 40 prompts at every size, the barrier grows from cycle 2 to cycle 6 in 34 of 36, 35 of 37, 33 of 33 and 33 of 33 loops from the smallest to the largest model, and the saturating law fits best for 22 to 26 loops per model. The saturated barrier rises with size, with medians 3.0, 3.6, 4.5 and 4.8 from 70M to 1B, and so does the temperature needed to leave a loop. The median $T_{1/2}$ for a horizon of 16 cycles is 0.43, 0.48, 0.59 and 0.64, and for 256 cycles it is 0.36, 0.42, 0.52 and 0.53. Over a 14-fold increase in parameters the locking temperature therefore rises by about half. Free generations of the 410M and 1B models (eight temperatures from 0.3 to 1.2, three samples per prompt) show a wider coherent window than at 160M. The coherent fraction exceeds one half only at $T=0.8$ for Pythia-160M, and from $T=0.7$ to $T=0.9$ for 410M and 1B, with peaks of 81\% and 82\%. Within this family, larger models produce coherent text over a wider band of temperatures, while the exact loops they do enter are harder to leave.

\begin{figure}[t]
\centering
\includegraphics[width=\textwidth]{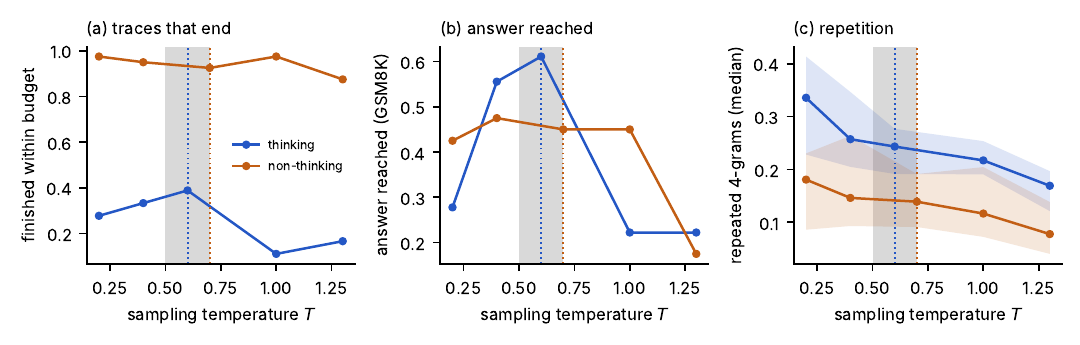}
\caption{Experiment 3, reasoning model. Qwen3-0.6B on GSM8K test problems with plain temperature sampling, with thinking (18 problems, 1024-token budget) and without (20 problems, two samples each, 640 tokens). (a) Fraction of traces that end inside the budget. (b) Fraction whose last number equals the answer. (c) Median fraction of repeated 4-grams (band: interquartile range). Gray band: the range recommended for DeepSeek-R1; dotted lines: the Qwen3 recommendations for thinking (0.6) and non-thinking (0.7) mode. No trace ends in an exact loop.}
\label{fig:exp3f}
\end{figure}

\paragraph{Detail: Reasoning models and the recommended temperature.} Vendors of reasoning models prescribe a narrow sampling range. The DeepSeek-R1 model card asks for a temperature between 0.5 and 0.7 ``to prevent endless repetitions or incoherent outputs'' \citep{deepseekr1card,guo2025deepseekr1}, and the Qwen3 card recommends $T=0.6$ with thinking, $T=0.7$ without, and no greedy decoding \citep{qwen3card,yang2025qwen3}. We asked whether these values mark the same window between two failures. Qwen3-0.6B solved the first 18 problems of the GSM8K test set \citep{cobbe2021training} in thinking mode with plain temperature sampling and a budget of 1024 tokens, one sample per problem and temperature (Figure~\ref{fig:exp3f}). The fraction of traces whose last number equals the answer is 0.28 at $T=0.2$, 0.56 at 0.4, 0.61 at 0.6, and 0.22 at both 1.0 and 1.3, and the fraction that also ends inside the budget follows the same shape (0.22, 0.28, 0.33, 0.06, 0.11). The best temperature on our grid is therefore the vendor's value, although with 18 problems per temperature the difference between $T=0.6$ and $T=0.2$ is at the edge of significance (Fisher exact test, p-value 0.09; 0.04 against $T=1.0$). The two failures are not those of the loops above. No trace ends in an exact loop. At $T=0.2$ the trace keeps returning to the same step with small rephrasings (``Wait, let me check again''), its median fraction of repeated 4-grams is 0.34 against 0.24 at $T=0.6$, and 13 of 18 traces run out of budget. At $T=1.3$ stray tokens from other scripts and broken phrases appear. In the language of Section~\ref{sec:regimes}, the low-temperature failure of this reasoning model is a recurrent orbit, not a periodic one, so recurrence quantification \citep{eckmann1987recurrence,marwan2007recurrence} and not the exact escape probabilities is the natural tool (\op{5}). Without thinking, the same model solves 0.42 to 0.48 of the first 20 problems at every temperature from 0.2 to 1.0, falling to 0.18 at 1.3, and never loops, so the repetitive edge appears only in the long thinking traces, consistent with the horizon dependence of the loops above. A single 0.6B model with a short budget tests the shape of the window and cannot confirm the vendors' exact numbers.

\begin{figure}[h]
\centering
\includegraphics[width=0.75\textwidth]{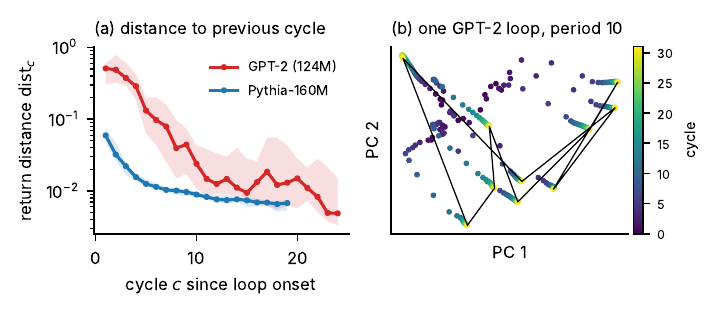}
\caption{Experiment 3, loops as orbits. (a) Return distance $\mathrm{dist}_c$ (mean distance between the final hidden state and its value one period earlier, relative to the mean state norm) against the cycle; line: median over loops, band: interquartile range. (b) First two principal components of the final hidden states (each dimension standardized) along one GPT-2 loop of period 10 (``He was a man of great wealth and power.''), colored by cycle. Each of the ten states of the cycle converges along its own track; black: the last cycle.}
\label{fig:exp3c}
\end{figure}

\paragraph{Loops converge onto periodic orbits in hidden-state space.} Along each greedy loop we also record the final-layer hidden state $x^L_n$ of the last token and compute the return distance $\mathrm{dist}_c$, the mean of $\|x^L_n-x^L_{n-k}\|$ over the $k$ steps of cycle $c$, divided by the mean norm of the states. If the loop is an attracting periodic orbit of the closed-loop map, $\mathrm{dist}_c\to0$. The return distance falls between cycle 2 and cycle 6 in 33 of 36 GPT-2 loops and 35 of 35 Pythia-160M loops. Its median falls from 0.51 in the first cycle to 0.013 in the twelfth in GPT-2, and from 0.060 to 0.008 in Pythia-160M (Figure~\ref{fig:exp3c}). It does not reach zero. Over the later cycles it stays near 0.01 in both models, because the context keeps growing, so the closed-loop map is not autonomous and the orbit keeps drifting slowly.

\emph{Loop detection.} From each prompt we decode greedily for 320 tokens and store the full logit vector at each step. A loop of period $k\le64$ is detected when the last $3k$ tokens are $k$-periodic, and its onset is the earliest index from which the sequence stays $k$-periodic. \emph{Stay probabilities.} Along the loop the context is deterministic, so the probability \eqref{eq:stay} of producing the loop token at temperature $T$ follows from the stored logits, for 30 temperatures spaced geometrically between $0.15$ and $2$. From these we compute, for each cycle $c$, the barrier $\Delta_c$ (smallest gap between the loop token's logit and any other logit during the cycle) and the escape probability $r_c(T)=1-\prod_{t\in c}p_t(T)$. \emph{Arrhenius fit.} For the first cycle after two completed cycles of burn-in, we fit $\log r$ against $1/T$ by least squares over the temperatures where $10^{-9}<r<10^{-2}$, and compare the negative slope with $\Delta_{\min}$ of that cycle. \emph{Monte Carlo check.} For the first four loops we sample 128 continuations at $T\in\{0.6,1.0\}$ (nearest grid temperature) from the context after burn-in, record the first step whose token differs from the loop, and compare the fraction still in the loop after 1, 2 and 4 cycles with the exact survival probability. \emph{Free sampling and coherence.} For each of 12 temperatures between 0.2 and 2 we sample three continuations of 200 tokens from every prompt with plain temperature sampling (no top-k or top-p), record whether each ends in a loop with at least three repetitions, the fraction of repeated 4-grams, and the mean entropy of the sampling distribution along the path. Each continuation is then scored by Qwen2.5-0.5B \citep{qwen2025qwen25} conditioned on its prompt, as the mean negative log-likelihood per reference token. The human baseline applies the same score to 200 passages of OpenWebText (the \texttt{stas/openwebtext-10k} sample), each split into its first 16 GPT-2 tokens as prompt and the next 200 as continuation. \emph{Growth laws.} For each loop with at least five cycles after the first, we fit the barriers $\Delta_c$, $c\ge2$, with $\Delta_1+\gamma\ln c$ and $\Delta_1+b(c-1)$ by least squares and with $\Delta_\infty-(\Delta_\infty-\Delta_2)e^{-(c-2)/c_\ast}$ by nonlinear least squares, and pick the law with the smallest corrected Akaike criterion. The finite-horizon locking temperature $T_{1/2}(N)$ of a loop is the temperature, interpolated in $\log T$ on the 30-point grid, at which the exact probability of surviving $N$ cycles after burn-in equals one half; beyond the stored cycles the escape probability of the last stored cycle is repeated. We fit \eqref{eq:thalf} to the medians for $N\ge4$. \emph{Escape control.} The controller of \op{12} samples at a base temperature $T_0$. When the generated tokens end in at least $\max(2,\lceil 8/k\rceil)$ repetitions of a block of $k\le64$ tokens, it computes the escape probability of the last cycle on a grid of temperatures from the 64 largest logits of each step, raises the temperature for the next $k$ steps to the smallest value with escape probability $r_\ast$ (capped at 1.5), and then returns to $T_0$. On the 40 prompts with three samples each, $T_0=0.3$ and $r_\ast=0.9$ give 6\% loops and 94\% repetitive outputs (repeated 4-grams above the human 95th percentile), against 68\% and 32\% at constant $T=0.3$. A repetition penalty of 1.2 \citep{keskar2019ctrl} removes loops and repetition but makes 73\% of the outputs incoherent by our threshold. \emph{Model size.} For Pythia-70M, 410M and 1B we run the greedy loop analysis without the Monte Carlo check, Pythia-1B in two halves of 20 prompts to fit in memory, and the free generations of 410M and 1B at eight temperatures in batches of 10 prompts, scoring them with the reference model in a separate pass. \emph{Return distance.} We repeat the greedy runs while storing the final-layer hidden state (after the last layer norm) at each step. Figure~\ref{fig:exp3c} shows the median return distance and a projection of one GPT-2 loop on its first two principal components.

\emph{Reasoning model.} We use Qwen3-0.6B in float32 with its own chat template, switching thinking on or off through the template flag, and append to each GSM8K question the instruction to reason step by step and put the final answer in \verb|\boxed{}|. Sampling is plain temperature sampling with top-k and top-p disabled, in batches of 6 (thinking) or 10 (non-thinking) left-padded prompts, with a fixed seed per temperature, sample and batch. The answer is the content of the last \verb|\boxed{}| if present and otherwise the last number in the trace, compared numerically with the reference answer. A trace counts as finished when it emits the end-of-turn token inside the budget. Loops are detected as in Experiment~3 with periods up to 256 tokens, and a loop must repeat over at least 30 tokens, which rules out runs such as the zeros of a number cut by the budget. The thinking runs take about 28 minutes per temperature on two CPU cores.

\emph{Paired figures and regime atlas.} Figures~\ref{fig:pair_training}, \ref{fig:pair_escape}, \ref{fig:pair_depth} and~\ref{fig:catapult} are produced by \texttt{paper/fig\_pairs.py}, and Figure~\ref{fig:atlas} by \texttt{exp3\_escape/atlas.py}. The Duffing oscillator is integrated with the Euler-Maruyama scheme (step 0.025) after a transient of 100 time units and sampled once per time unit. For the recurrence plots the state is standardized and $\epsilon_R$ is chosen so that 5\% of all pairs recur; DET counts diagonal lines of length at least 2 and excludes the main diagonal. The spectral entropy is the Shannon entropy of the normalized Welch spectrum (segments of 256 samples, zero frequency excluded) divided by its maximum. The statistics in Section~\ref{sec:tools} (\texttt{exp3\_escape/atlas\_stats.py}) repeat these measurements for noise seeds 0 to 9 and for the first 10 prompts of Experiment~3; at $T=0.3$, 9 of the 10 GPT-2 runs end in an exact loop.

\emph{Controllers for clock 3.} Rollback and kick (\texttt{exp3\_escape/rollback.py}) samples at a base temperature $T_0$; when the generated tokens end in a loop it deletes the periodic tail back to its first token, forbids that token at that position and resamples there at $T_0$. The feedback controller (\texttt{exp3\_escape/regime\_control.py}) updates the temperature after every token from the last 32 tokens, $T\leftarrow\mathrm{clip}\big(T+0.05(\mathrm{rep}-0.02)-0.01\max(0,\mathcal{I}-3),\,0.4,\,1.1\big)$, where $\mathrm{rep}$ is the fraction of 4-grams in the window that already occurred earlier in the generation and $\mathcal{I}$ is the mean surprisal of the window at $T=1$ in nats. The gains were set by hand on three of the 40 prompts; without these three prompts the coherent fractions are 21\%, 64\% and 82\%. Both use the 40 prompts, three samples and 200 tokens of Experiment~3b, and are scored with the same reference model and thresholds, so the constant-temperature rows of Experiment~3b are the baseline.

\section{Further related work}
\label{app:related}

This appendix collects review material on each clock that supports Sections~\ref{sec:dictionary} and~\ref{sec:reduced} but is not needed to follow the main argument.

\subsection{Clock 1: Attention models closer to real decoders}

Later work moved the particle model of Section~\ref{sec:particles} closer to real decoders. With causal masking the particle system is no longer a mean-field gradient flow, but clustering results still hold, and the first token, which attends only to itself, plays a special role \citep{karagodin2024clustering}. \citet{wu2024role} showed that the attention mask and layer normalization decide whether collapse happens at all. A mean-field limit with a Vlasov-type ``Transformer PDE'' covers several attention variants \citep{castin2025unified}. Feed-forward layers can be read as a control that steers the particles \citep{maranzatto2026role}. Finite-width transformers with MLP blocks converge to a stochastic interacting particle system, in which noise can synchronize the tokens \citep{agazzi2026stochastic}. For symmetric weights a spectral mode-selection mechanism decides the long-time state \citep{kuehn2026spectral}, and in two dimensions deep linear transformers reduce to Kuramoto-type models with a rich set of behaviors \citep{li2026diverse}. The kinetic theory of causal attention has also been used to explain why models use the middle of long contexts poorly \citep{duerinckx2026kinetic}, the lost-in-the-middle effect of \citet{liu2023lost}.

Further measurements of the depth trajectory include the rise and fall of the intrinsic dimension of hidden representations across layers \citep{valeriani2023geometry}, smooth changes of individual units from layer to layer seen with tools from neural population dynamics \citep{fernando2025transformer}, full Jacobian eigendecompositions of each layer in production-scale models \citep{fernando2026dynamics}, a theoretical analysis of methods that read features off the layer Jacobian \citep{ostrow2026mathematical}, and looped models whose repeated block makes the iteration explicit and whose stability can be studied and scaled as such \citep{prairie2026parcae}.

\subsection{Clock 1: Lessons from recurrent networks}

Recurrent networks are the older case in which clocks~1 and~3 coincide: the same map is applied at every time step. Their analysis set many of the patterns used later for transformers. Fixed points and slow points of trained networks explain how they compute \citep{sussillo2013opening}, and sentiment classifiers turn out to integrate evidence along a line attractor \citep{maheswaranathan2019reverse}. On the training side, exploding gradients were linked to bifurcations of the network dynamics \citep{pascanu2012difficulty,ribeiro2019exploding}, and networks trained to reproduce chaotic dynamics must have diverging gradients \citep{mikhaeil2021difficulty}. Lyapunov spectra quantify how much information a recurrent network keeps \citep{engelken2020lyapunov,vogt2020lyapunov}.

\subsection{Clocks 1 and 3: Explicit dynamical models of sequences}

State-space models write the sequence dynamics in clock~3 explicitly, as a linear dynamical system in a hidden state, with eigenvalues placed to control memory and stability \citep{gu2021efficiently,orvieto2023resurrecting}, or with input-dependent transitions \citep{gu2023mamba}. Linear attention has the same structure: it is a recurrent update of a matrix-valued state \citep{katharopoulos2020transformers,schlag2021linear}. Recurrent networks built from coupled oscillators or constrained by Lipschitz bounds use the same design logic \citep{rusch2020coupled,erichson2020lipschitz}. In these models the attractor and stability questions of this paper are explicit, since the eigenvalues of the state transition are parameters of the architecture.

\subsection{Clock 2: Theory of the edge of stability}

\citet{marion2026edge} proposes a three-variable flow for the center, direction and amplitude of the oscillation at the edge of stability. \citet{song2023trajectory} prove that in simple models trajectories align on a bifurcation diagram, \citet{gan2026bifurcation} treats the onset of oscillation as a flip bifurcation of the normal dynamics near a manifold of minimizers, and \citet{mulayoff2026stability} give an exact criterion, in terms of third and fourth derivatives, for when oscillations near a minimum are stable. Why the sharpness rises in the first place, progressive sharpening, is proved only in reduced models \citep{agarwala2022secondorder}, and the route from the edge to chaos at larger steps has been mapped in a two-layer linear model \citep{kalra2023universal}.

\subsection{Clock 2: Moving and controlling the stability boundary}

Mini-batch sampling moves the stability boundary of gradient descent even when gradients are unbiased, as shown exactly for linear self-attention \citep{balasubramanian2026unbiased}, and curvature control has been used to keep large transformers away from the boundary \citep{ramasinghe2026taming,gilmer2021loss}.

\subsection{Clock 2: Plateaus, grokking and emergence}

Noise or small perturbations let gradient methods leave strict saddles efficiently \citep{ge2015escaping,jin2017escape}, but the time spent near a saddle can still dominate training. Spectral bias and the frequency principle describe an ordering similar to the plateaus of deep linear networks, with low-frequency components learned first \citep{rahaman2018spectral,xu2019frequency}, and learning can itself drive a recurrent network from chaotic to stable activity \citep{vaidya2026learninginduced}. Explanations of grokking include an effective theory of representation learning with a phase diagram \citep{liu2022grokking}, a dependence on weight norm \citep{liu2022omnigrok}, a transition from lazy to rich training \citep{kumar2023grokking} and two phases of implicit bias \citep{lyu2023dichotomy}; with adaptive optimizers, grokking coincides with cyclic phase changes and loss spikes \citep{thilak2022slingshot}. In language models, syntactic attention structure appears abruptly and is followed by a drop in loss \citep{chen2023sudden}. In-context learning can emerge and later fade while training continues \citep{singh2023transient}; minimal models explain the abruptness \citep{reddy2023mechanistic}, and Bayesian models of attention predict phase transitions in the emergence of copy heads \citep{lavie2026phase}. Singular learning theory relates these stages to changes in the degeneracy of the loss landscape \citep{hoogland2024loss}, and one model attributes emergence to discrete skills that are learned in order of frequency \citep{michaud2023quantization}.

\subsection{Clock 2: Slow phases, landscape geometry and conserved quantities}

Pretraining curves typically show a fast initial drop followed by a long slow phase. \citet{zhang2026stability} observe that the normalized singular value spectrum of the weight matrices stabilizes early and that this stabilization coincides with the slow phase. The river-valley picture explains why a constant learning rate followed by a fast decay works: the iterate moves along a valley floor while oscillating across it \citep{wen2024warmupstabledecay}. The Hessian has a bulk of near-zero eigenvalues and a few outliers \citep{sagun2017empirical,ghorbani2019investigation}, and the gradient lies mostly in the subspace of the outliers \citep{gurari2018gradient}. Two copies of a run that are split at step $k$ and then see different minibatch noise end in the same linearly connected region only when $k$ is past a short early phase; before that, the trajectory is sensitive to small perturbations \citep{frankle2019linear}. Continuous symmetries of the architecture imply conservation laws for gradient flow, which finite steps and noise break in predictable ways \citep{kunin2020neural}. At the end of classification training, features converge to a highly symmetric configuration \citep{papyan2020prevalence}.

Infinite-width limits give two simpler dynamical systems. In the lazy limit, training is linear in function space \citep{jacot2018neural}. In the mean-field limit, training two-layer networks is a Wasserstein gradient flow of particles \citep{mei2018mean,chizat2018global}, and parametrizations exist that keep feature learning at infinite width \citep{yang2020feature}. Scaling laws describe the slow phase at the level of loss averaged over the whole run \citep{kaplan2020scaling,hoffmann2022training}. Koopman operator methods have been used to speed up training and to explain magnitude pruning \citep{dogra2020optimizing,redman2021operator}.

\subsection{Clock 3: Outer loops, when generation feeds training}

The slow-fast separation of Section~\ref{sec:maps} breaks when generated text is used for training. Training repeatedly on a model's own outputs makes the tails of the distribution disappear \citep{shumailov2023curse,alemohammad2023selfconsuming}; whether the process stays near the original model depends on the fraction of real data in each round \citep{bertrand2023stability}. Reinforcement learning from human feedback reduces output diversity \citep{kirk2023effects}, which can be read as a contraction of the set of trajectories that clock~3 visits. These are dynamical systems over model versions, a fourth and much slower clock.

\subsection{Clock 3: Language models as models of dynamical systems}

A different line of work asks whether language models can forecast dynamical systems placed in their context. Pretrained models forecast time series zero-shot \citep{gruver2023large} and extrapolate trajectories of physical systems with accuracy that improves with context length \citep{liu2024llms}. Chaotic systems serve as a benchmark for such forecasters \citep{gilpin2021chaos,gilpin2023model,pathak2018model}, and training on data from cellular automata at the edge of chaos produces models with better downstream reasoning \citep{zhang2024intelligence}. These results are about the content a model processes, not about the model as a dynamical system, but they share tools with this paper.

\end{document}